\documentclass{article}
\usepackage{amsfonts, amsmath, booktabs, microtype, multirow, enumitem, graphicx}
\usepackage[english]{babel}
\usepackage[authoryear]{natbib}
\usepackage[preprint]{neurips_2020}
\setlist{leftmargin=5mm}

\usepackage{listings}
\usepackage[table]{xcolor}
\usepackage{hyperref}
\usepackage{float}
\usepackage{xspace}

\hypersetup{
    colorlinks,
    linkcolor={blue!50!black},
    citecolor={blue!50!black},
    urlcolor={blue!50!black}
}

\newcommand{\dtr}{{d_\mathrm{tr}}}
\newcommand{\dte}{{d_\mathrm{te}}}
\newcommand{\domainbed}{\textsc{DomainBed}\xspace}

\colorlet{alternateRowColor}{teal!5}

\newcommand{\coloredBelowRuleSep}[1]{
    \arrayrulecolor{#1}
    \specialrule{\belowrulesep}{0pt}{0pt}
    \arrayrulecolor{black}
}

\newcommand{\coloredMidrule}[2]{
    \arrayrulecolor{#1}
    \specialrule{\aboverulesep}{0pt}{0pt}
    \arrayrulecolor{black}
    \specialrule{\lightrulewidth}{0pt}{0pt}
    \coloredBelowRuleSep{#2}
}

\newcommand{\coloredBottomrule}[1]{
    \arrayrulecolor{#1}
    \specialrule{\aboverulesep}{0pt}{0pt}
    \arrayrulecolor{black}
    \specialrule{\heavyrulewidth}{0pt}{0pt}
    \coloredBelowRuleSep{white}
}

\newtheorem{recommendation}{Recommendation}

\title{In Search of Lost Domain Generalization}
\author{%
    Ishaan Gulrajani and David Lopez-Paz\thanks{Alphabetical order, equal contribution.}\\
    Facebook AI Research\\
    \texttt{igul222@gmail.com, dlp@fb.com}
}

\begin{document}

\maketitle

\begin{abstract}
The goal of domain generalization algorithms is to predict well on distributions different from those seen during training.
While a myriad of domain generalization algorithms exist, inconsistencies in experimental conditions---datasets, architectures, and model selection criteria---render fair and realistic comparisons difficult.
In this paper, we are interested in understanding how useful domain generalization algorithms are in realistic settings.
As a first step, we realize that model selection is non-trivial for domain generalization tasks.
Contrary to prior work, we argue that domain generalization algorithms without a model selection strategy should be regarded as incomplete.
Next, we implement \domainbed, a testbed for domain generalization including seven multi-domain datasets, nine baseline algorithms, and three model selection criteria.
We conduct extensive experiments using \domainbed and find that, when carefully implemented, empirical risk minimization shows state-of-the-art performance across all datasets.
Looking forward, we hope that the release of \domainbed, along with contributions from fellow researchers, will streamline reproducible and rigorous research in domain generalization.
\end{abstract}

\section{Introduction}

Machine learning systems often fail to \emph{generalize out-of-distribution}, crashing in spectacular ways when tested outside the domain of training examples \citep{torralba2011unbiased}.
The overreliance of learning systems on the training distribution manifests widely.
For instance, self-driving car systems struggle to perform under conditions different to those of training, including variations in light \citep{dai2018dark}, weather \citep{volk2019towards}, and object poses \citep{alcorn2019strike}.
As another example, systems trained on medical data collected in one hospital do not generalize to other health centers \citep{castro2019causality, albadawy2018deep, perone2019unsupervised, mittech}.
\citet{arjovsky2019invariant} suggest that failing to generalize out-of-distribution is failing to capture the causal factors of variation in data, clinging instead to easier-to-fit spurious correlations, which are prone to change from training to testing domains.
Examples of spurious correlations commonly absorbed by learning machines include racial biases \citep{stock2018convnets}, texture statistics \citep{geirhos2018ImageNet}, and object backgrounds \citep{beery2018recognition}.
Alas, the capricious behaviour of machine learning systems out-of-distribution is a roadblock to their deployment in critical applications.

Aware of this problem, the research community has spent significant effort during the last decade to develop algorithms able to generalize out-of-distribution.
In particular, the literature in \emph{domain generalization} assumes access to multiple datasets during training, each of them containing examples about the same task, but collected under a different domain or environment \citep{blanchard2011generalizing, muandet2013domain}. 
The goal of domain generalization algorithms is to incorporate the invariances across these training datasets into a classifier, in hopes that such invariances also hold in novel test domains.
Different domain generalization solutions assume different types of invariances and propose algorithms to estimate them from data.

Despite the enormous importance of domain generalization, the literature is scattered: a plethora of different algorithms appear yearly, and these are evaluated under different datasets and model selection criteria.
Borrowing from the success of standard computer vision benchmarks such as ImageNet \citep{russakovsky2015ImageNet}, the purpose of this work is to perform a standardized, rigorous comparison of domain generalization algorithms.
In particular, we ask: how useful are domain generalization algorithms in realistic settings?
Towards answering this question, we first study model selection criteria for domain generalization methods, resulting in the recommendation: 
\begin{center}
    \textit{A domain generalization algorithm should be responsible for specifying a model selection method.}
\end{center}
We then carefully implement nine domain generalization algorithms on seven multi-domain datasets and three model selection criteria, leading us to the conclusion reflected in Tables~\ref{table:summary} and \ref{table:main_results}:
\begin{center}
    \textit{When equipped with modern neural network architectures and data augmentation techniques, empirical risk minimization achieves state-of-the-art performance in domain generalization.}
\end{center}

\begin{table}
    \caption{State-of-the-art domain generalization for typical datasets and their domains.
    Our implementation of Empirical Risk Minimization (ERM) outperforms previous literature.}
    \begin{center}
    \begin{tabular}{lccccccc}
        \toprule
        \textbf{Dataset / algorithm} & \multicolumn{7}{c}{\textbf{Out-of-distribution accuracy (by domain)}} \\
        \coloredMidrule{white}{alternateRowColor}
        \rowcolor{alternateRowColor}
        Rotated MNIST & $0^{\circ}$ & $15^{\circ}$ & $30^{\circ}$ & $45^{\circ}$ & $60^{\circ}$ & $75^{\circ}$ & Average \\
        \coloredBelowRuleSep{white}
        \quad \cite{ilse2019diva}        & 93.5 & 99.3 & 99.1 & 99.2 & 99.3 & 93.0 & 97.2 \\
        \quad Our ERM                    & 95.6 & 99.0 & 98.9 & 99.1 & 99.0 & 96.7 & \textbf{98.0} \\
        \coloredMidrule{white}{alternateRowColor}
        \rowcolor{alternateRowColor}
        PACS & A & C & P & S &  &  & Average \\
        \coloredBelowRuleSep{white}
        \quad \cite{asadi2019towards}    & 83.0 & 79.4 & 96.8 & 78.6 &      &      & 84.5 \\
        \quad Our ERM                    & 88.1 & 78.0 & 97.8 & 79.1 &      &      & \textbf{85.7} \\
        \coloredMidrule{white}{alternateRowColor}
        \rowcolor{alternateRowColor}
        VLCS & C & L & S & V &  &  & Average \\
        \coloredBelowRuleSep{white}
        \quad \cite{albuquerque2019a}    & 95.5 & 67.6 & 69.4 & 71.1 &      &      & 75.9 \\
        \quad Our ERM                    & 97.6 & 63.3 & 72.2 & 76.4 &      &      & \textbf{77.4} \\
        \coloredMidrule{white}{alternateRowColor}
        \rowcolor{alternateRowColor}
        Office-Home & A & C & P & R &  &  & Average \\
        \coloredBelowRuleSep{white}
        \quad \cite{zhou2020deep}        & 59.2 & 52.3 & 74.6 & 76.0 &      &      & 65.5 \\
        \quad Our ERM                    & 62.7 & 53.4 & 76.5 & 77.3 &      &      & \textbf{67.5} \\
        \bottomrule
    \end{tabular}
    \end{center}
    \label{table:summary}
\end{table}

As a result of our research, we release \domainbed, a framework to streamline rigorous and reproducible experimentation in domain generalization.
Using \domainbed, adding a new algorithm or dataset is a matter of a few lines of code; a single command runs all the experiments, performs all the model selections, and auto-generates all the tables included in this work.
Moreover, our motivation is to keep \domainbed alive, welcoming pull requests from our fellow colleagues to update the available algorithms, datasets, model selection criteria, and result tables. 

Section~\ref{sec:setup} kicks off our exposition with a review of the domain generalization setup.
Section~\ref{sec:hparam_selection} discusses the difficulties of model selection in domain generalization and makes recommendations for a path forward.
Section~\ref{sec:benchmark} introduces \domainbed, describing the algorithms and datasets contained in the initial release.
Section~\ref{sec:experiments} discusses the experimental results of running the entire \domainbed suite; these illustrate the strength of ERM and the importance of model selection criteria.
Finally, Section~\ref{sec:outlook} offers our view on future research directions in domain generalization.
Our Appendices review one hundred articles spanning a decade of research in this topic, collecting the experimental performance of over thirty published algorithms.

\section{The problem of domain generalization}
\label{sec:setup}

The goal of supervised learning is to predict values $y \in \mathcal{Y}$ of a target random variable $Y$, given values $x \in \mathcal{X}$ of an input random variable $X$.
Predictions $\hat{y}$ about $x$ originate from a predictor $f : \mathcal{X} \to \mathcal{Y}$, such that $\hat{y} = f(x)$.
We often decompose predictors as $f = w \circ \phi$, where we call $\phi : \mathcal{X} \to \mathcal{H}$ the featurizer, and $w : \mathcal{H} \to \mathcal{Y}$ the classifier.
Our main tool to solve the prediction task is \emph{the} training dataset $D = \{(x_i, y_i)\}_{i=1}^n$, which contains identically and independently distributed (iid) examples from the joint probability distribution $P(X, Y)$.
Given a loss function $\ell : \mathcal{Y} \times \mathcal{Y} \to [0, \infty)$ measuring the prediction error at one example, we often cast supervised learning as finding a predictor minimizing the population risk $\mathbb{E}_{(x, y) \sim P}[\ell(f(x), y)]$.
Since we only have access to the data distribution $P(X, Y)$ via the dataset $D$, we instead choose a predictor minimizing the \emph{empirical} risk $\frac{1}{n} \sum_{i=1}^n \ell(f(x_i), y_i)$
\citep{vapnik1998statistical}.

The rest of this paper studies the problem of domain generalization, an extension of supervised learning where training datasets from multiple domains (or environments) are available to train our predictor \citep{blanchard2011generalizing}. 
More specifically, we characterize each domain $d$ by a dataset $D^d = \{(x_i^d, y_i^d)\}_{d=1}^{n_d}$ containing iid examples from some probability distribution $P(X^d, Y^d)$, for all training domains $d \in \{1, \ldots, \dtr\}$.
The goal of domain generalization is \emph{out-of-distribution generalization}: learning a predictor able to perform well at some unseen test domain $\dte = \dtr + 1$.
Since no data about the test domain is available during training, we must assume the existence of some statistical invariances across training and testing domains in order to incorporate such invariances (but nothing else) into our predictor.
The type of invariance assumed, as well as how to estimate it from the training datasets, varies between domain generalization algorithms.

Domain generalization differs from unsupervised domain adaptation. In the latter, it is assumed that unlabeled data from the test domain is available during training \citep{pan2009survey, patel2015visual, wilson2018survey}.
Table~\ref{table:paradigms} compares different machine learning setups to highlight the nature of domain generalization problems.
The causality literature refers to domain generalization as \emph{learning from multiple environments} \citep{peters2016causal, arjovsky2019invariant}.
Although challenging, domain generalization is the best approximation to real prediction problems, where unforeseen distributional discrepancies between training and testing data are surely expected.

\begin{table}
    \caption{Learning setups.
    $L^d$ and $U^d$ denote the labeled and unlabeled distributions from domain $d$.}
    \begin{center}
    \begin{tabular}{lll}
        \toprule
        \textbf{Setup} & \textbf{Training inputs} & \textbf{Test inputs} \\
        \coloredMidrule{white}{alternateRowColor}
        \rowcolor{alternateRowColor}
        Generative learning        & $U^1$ & $\emptyset$ \\
        Unsupervised learning      & $U^1$ & $U^1$ \\
        \rowcolor{alternateRowColor}
        Supervised learning        & $L^1$ & $U^1$ \\
        Semi-supervised learning   & $L^1, U^1$ & $U^1$ \\
        \rowcolor{alternateRowColor}
        Multitask learning         & $L^1, \ldots, L^\dtr$ & $U^1, \ldots, U^\dtr$ \\
        Continual (or lifelong) learning         & $L^1, \ldots, L^\infty$ & $U^1, \ldots, U^\infty$\\
        \rowcolor{alternateRowColor}
        Domain adaptation          & $L^1, \ldots, L^\dtr, U^{\dtr + 1}$ & $U^{\dtr + 1}$ \\
        Transfer learning          & $U^1, \ldots, U^\dtr, L^{\dtr + 1}$ & $U^{\dtr + 1}$ \\
        \rowcolor{alternateRowColor}
        \textbf{Domain generalization}      & $L^1, \ldots, L^\dtr$ & $U^{\dtr + 1}$\\
        \coloredBottomrule{alternateRowColor}
    \end{tabular}
    \end{center}
    \label{table:paradigms}
\end{table}

\section{Model selection as part of the learning problem}
\label{sec:hparam_selection}

Here we discuss issues surrounding model selection (choosing hyperparameters, training checkpoints, architecture variants) in domain generalization and make specific recommendations for a path forward.
Because we lack access to a validation set identically distributed to the test data, model selection in domain generalization is not as straightforward as in supervised learning.
Some works adopt heuristic strategies whose behavior is not well-studied, while others simply omit a description of how to choose hyperparameters.
This leaves open the possibility that hyperparameters were chosen using the test data, which is not methodologically sound.
Differences in results arising from inconsistent tuning practices may be misattributed to the algorithms under study, complicating fair assessments.

We believe that much of the confusion surrounding model selection in domain generalization arises from treating it as a question of experimental design.
In reality, selecting hyperparameters is a learning problem at least as hard as fitting the model (inasmuch as we may interpret any model parameter as a hyperparameter).
Like all learning problems, model selection requires assumptions about how the test data relates to the training data.
Different domain generalization algorithms make different assumptions, and it is not clear \textit{a priori} what assumptions are correct, or how these assumptions influence the model selection criterion. 
Indeed, choosing reasonable assumptions is at the heart of domain generalization research.
Therefore, a domain generalization algorithm without a strategy to choose its hyperparameters remains incomplete.

\begin{recommendation}
A domain generalization algorithm should be responsible for specifying a model selection method.
\end{recommendation}

While algorithms without well-justified model selection methods are incomplete, they may be useful as stepping-stones in a research agenda.
In this case, instead of using an ad-hoc model selection method, we can evaluate incomplete algorithms by considering an \emph{oracle model selection method}, where we select hyperparameters on the test domain.
Of course, it is important that we avoid invalid comparisons between oracle results and baselines tuned without an oracle method.
Also, unless we restrict access to the test domain data somehow, we risk obtaining meaningless results.
For instance, we could just train on such test domain data using supervised learning.

\begin{recommendation}
Researchers should disclaim any oracle-selection results as such and specify policies to limit access to the test domain.
\end{recommendation}

\subsection{Three model selection methods}

Having made broad recommendations, we review and justify three methods for model selection in domain generalization, often used but rarely discerned.

\paragraph{Training-domain validation set}

We split each training domain into training and validation subsets.
Then, we pool the validation subsets of each training domain to create an overall validation set.
Finally, we choose the model maximizing the accuracy on the overall validation set.

This strategy assumes that the training and test examples follow similar distributions.
For example, \citet{ben2010theory} bound the test domain error of a classifier by the training domain error, plus a divergence measure between the training and test domains.

\paragraph{Leave-one-domain-out cross-validation}

Given $\dtr$ training domains, we train $\dtr$ models with equal hyperparameters, each holding one of the training domains out.
We evaluate each model on its held-out domain, and average the accuracies of these models over their held-out domains.
Finally, we choose the model maximizing this average accuracy, re-trained on all $\dtr$ domains.

This strategy assumes that training and test domains are drawn from a \emph{meta-distribution} over domains, and that our goal is to maximize the expected performance under this meta-distribution.

\paragraph{Test-domain validation set (oracle)}

We choose the model maximizing the accuracy on a validation set that follows the distribution of the test domain.
Following our earlier recommendation to limit test domain access, we allow $20$ queries per algorithm (one query per choice of hyperparameters in our random search).
This means that we do not allow early stopping based on the validation set.
Instead, we train all models for the same fixed number of steps and consider only the final checkpoint.
Recall that we do not consider this a valid benchmarking methodology, since it requires access to the test domain.
Oracle-selection results can be either optimistic, because we access the test distribution, or pessimistic, because the query limit reduces the number of considered hyperparameter combinations.

As an alternative to limiting the number of queries, we could borrow tools from differential privacy, previously applied to enable multiple re-uses of validation sets in standard supervised learning \citep{dwork2015reusable}. 
In a nutshell, differential privacy tools add Laplace noise to the accuracy statistic of the algorithm before reporting it to the practitioner.

\subsection{Considerations from the literature}

Some references in prior work discuss additional strategies to choose hyperparemeters in domain generalization problems.
For instance, \citet[Appendix B.1]{krueger2020out} suggest choosing hyperparameters to maximize the performance across all domains of an external dataset.
The validity of this strategy depends on the relatedness between datasets.
\citet[Section 5.3.2]{albuquerque2019a} suggest performing model selection based on the loss function (which often incorporates an algorithm-specific regularizer), and \citet[Section 3]{d2018domain} derive an strategy specific to their algorithm.

\section{\domainbed: A PyTorch testbed for domain generalization}
\label{sec:benchmark}

At the heart of our large scale experimentation is \domainbed, a PyTorch \citep{paszke2019pytorch} testbed to streamline reproducible and rigorous research in domain generalization:
\begin{center}
    \url{https://github.com/facebookresearch/DomainBed} \textit{(coming soon)}
\end{center}

The initial release comprises nine algorithms, seven datasets, and three model selection methods (described in Section~\ref{sec:hparam_selection}), as well as the infrastructure to run all the experiments and generate all the \LaTeX{} tables below with a single command. 
\domainbed is a living project: we expect to update the above repository with new results, algorithms, and datasets. Contributions via pull requests from fellow researchers are welcome.
Adding a new algorithm or dataset to \domainbed is a matter of a few lines of code (see Appendix~\ref{sec:code_example} for an example).

\subsection{Datasets}
\label{sec:datasets}

\begin{table}
    \caption{Datasets included in \textsc{DomainBed}. For each dataset, we pick a single class and show illustrative images from each domain.}
    \begin{center}
    \begin{tabular}{lcccccc}
        \toprule
        \textbf{Dataset} & \multicolumn{6}{l}{\textbf{Domains}} \\
        \coloredMidrule{white}{alternateRowColor}
        \rowcolor{alternateRowColor}
        & \tiny{+90\%} & \tiny{+80\%} & \tiny{-90\%} & & & \\
        \rowcolor{alternateRowColor}
        Colored MNIST &
            \raisebox{-.5\height}{\includegraphics[width=25pt, height=25pt]{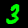}} &
            \raisebox{-.5\height}{\includegraphics[width=25pt, height=25pt]{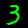}} &
            \raisebox{-.5\height}{\includegraphics[width=25pt, height=25pt]{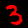}} & & &
            \\
        \rowcolor{alternateRowColor}
        & \multicolumn{6}{l}{\tiny{\emph{(degree of correlation between color and label)}}} \\
        & \tiny{0$^{\circ}$} & \tiny{15$^{\circ}$} & \tiny{30$^{\circ}$} & \tiny{45$^{\circ}$} & \tiny{60$^{\circ}$} & \tiny{75$^{\circ}$} \\
        Rotated MNIST &
            \raisebox{-.5\height}{\includegraphics[width=25pt, height=25pt]{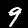}} &
            \raisebox{-.5\height}{\includegraphics[width=25pt, height=25pt]{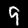}} &
            \raisebox{-.5\height}{\includegraphics[width=25pt, height=25pt]{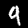}} &
            \raisebox{-.5\height}{\includegraphics[width=25pt, height=25pt]{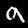}} &
            \raisebox{-.5\height}{\includegraphics[width=25pt, height=25pt]{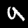}} &
            \raisebox{-.5\height}{\includegraphics[width=25pt, height=25pt]{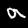}}
            \\
        \coloredBelowRuleSep{white}
        \coloredBelowRuleSep{white}
        \rowcolor{alternateRowColor}
        & \tiny{Caltech101} & \tiny{LabelMe} & \tiny{SUN09} & \tiny{VOC2007} & & \\
        \rowcolor{alternateRowColor}
        VLCS &
            \raisebox{-.5\height}{\includegraphics[width=25pt, height=25pt]{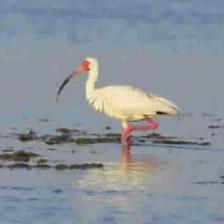}} &
            \raisebox{-.5\height}{\includegraphics[width=25pt, height=25pt]{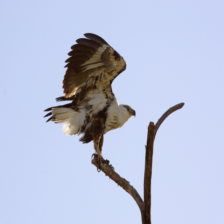}} &
            \raisebox{-.5\height}{\includegraphics[width=25pt, height=25pt]{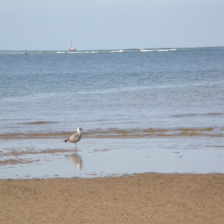}} &
            \raisebox{-.5\height}{\includegraphics[width=25pt, height=25pt]{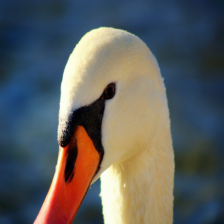}} & &
            \\
        \coloredBelowRuleSep{alternateRowColor}
        \coloredBelowRuleSep{alternateRowColor}
        & \tiny{Art} & \tiny{Cartoon} & \tiny{Photo} & \tiny{Sketch} & & \\
        PACS &
            \raisebox{-.5\height}{\includegraphics[width=25pt, height=25pt]{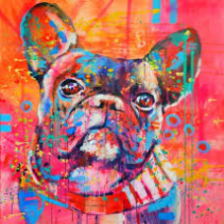}} &
            \raisebox{-.5\height}{\includegraphics[width=25pt, height=25pt]{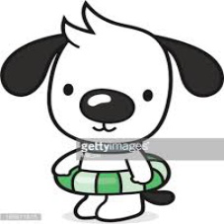}} &
            \raisebox{-.5\height}{\includegraphics[width=25pt, height=25pt]{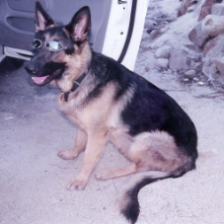}} &
            \raisebox{-.5\height}{\includegraphics[width=25pt, height=25pt]{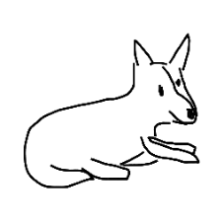}} & &
            \\
        \coloredBelowRuleSep{white}
        \coloredBelowRuleSep{white}
        \rowcolor{alternateRowColor}
        & \tiny{Art} & \tiny{Clipart} & \tiny{Product} & \tiny{Photo} & & \\
        \rowcolor{alternateRowColor}
        Office-Home &
            \raisebox{-.5\height}{\includegraphics[width=25pt, height=25pt]{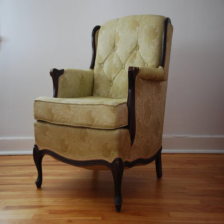}} &
            \raisebox{-.5\height}{\includegraphics[width=25pt, height=25pt]{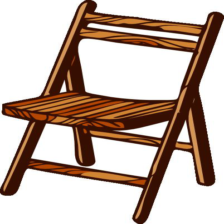}} &
            \raisebox{-.5\height}{\includegraphics[width=25pt, height=25pt]{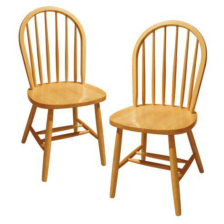}} &
            \raisebox{-.5\height}{\includegraphics[width=25pt, height=25pt]{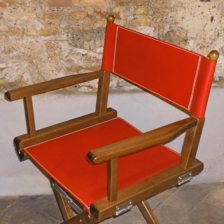}} & &
            \\
        \coloredBelowRuleSep{alternateRowColor}
        \coloredBelowRuleSep{alternateRowColor}
        & \tiny{L100} & \tiny{L38} & \tiny{L43} & \tiny{L46} & & \\
        Terra Incognita &
            \raisebox{-.5\height}{\includegraphics[width=25pt, height=25pt]{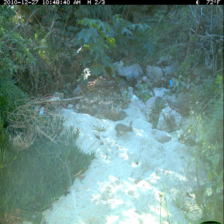}} &
            \raisebox{-.5\height}{\includegraphics[width=25pt, height=25pt]{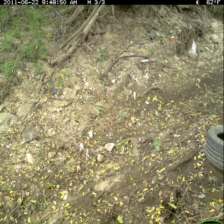}} &
            \raisebox{-.5\height}{\includegraphics[width=25pt, height=25pt]{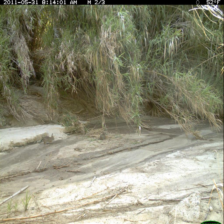}} &
            \raisebox{-.5\height}{\includegraphics[width=25pt, height=25pt]{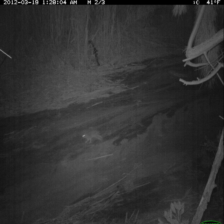}} &
            \\
        & \multicolumn{6}{l}{\tiny{\emph{(camera trap location)}}} \\
        \rowcolor{alternateRowColor}
        & \tiny{Clipart} & \tiny{Infographic} & \tiny{Painting} & \tiny{QuickDraw} & \tiny{Photo} & \tiny{Sketch} \\
        \rowcolor{alternateRowColor}
        DomainNet &
            \raisebox{-.5\height}{\includegraphics[width=25pt, height=25pt]{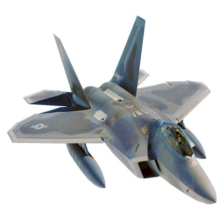}} &
            \raisebox{-.5\height}{\includegraphics[width=25pt, height=25pt]{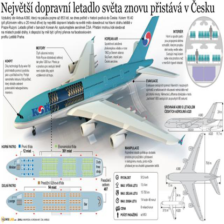}} &
            \raisebox{-.5\height}{\includegraphics[width=25pt, height=25pt]{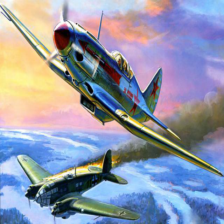}} &
            \raisebox{-.5\height}{\includegraphics[width=25pt, height=25pt]{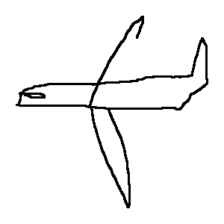}} &
            \raisebox{-.5\height}{\includegraphics[width=25pt, height=25pt]{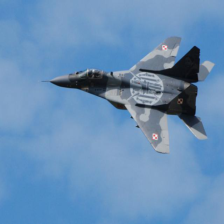}} &
            \raisebox{-.5\height}{\includegraphics[width=25pt, height=25pt]{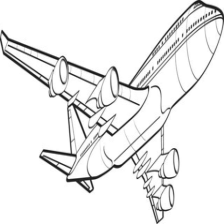}}
            \\
        \coloredBelowRuleSep{alternateRowColor}
        \coloredBottomrule{alternateRowColor}
    \end{tabular}
    \end{center}
    \label{table:datasets}
\end{table}

\domainbed includes downloaders and loaders for seven multi-domain image classification tasks:
Colored MNIST \citep{arjovsky2019invariant}, Rotated MNIST \citep{ghifary2015domain}, PACS \citep{Li_2017_ICCV}, VLCS \citep{fang2013unbiased}, Office-Home \citep{venkateswara2017Deep}, Terra Incognita \citep{beery2018recognition}, and DomainNet \citep{peng2019moment}.
We list and show example images from each dataset in Table~\ref{table:datasets}, and provide their full details in Appendix \ref{sec:dataset_details}.

The datasets differ in many ways but two are particularly important.
The first difference is between synthetic and real datasets.
In Rotated MNIST and Colored MNIST, domains are synthetically constructed such that we know what features will generalize \textit{a priori}, so using too much prior knowledge (e.g. by augmenting with rotations) is off-limits, whereas the other datasets contain domains arising from natural processes, making it sensible to use prior knowledge.
The second difference is about what changes across domains.
On one hand, in datasets other than Colored MNIST, the domain changes the distribution of images, but likely bears no information about the true image-to-label mapping.
On the other hand, in Colored MNIST, the domain influences the true image-to-label mapping, biasing algorithms that try to estimate this function directly.

\subsection{Algorithms}
\label{sec:algorithms}

The initial release of \domainbed includes implementations of nine baseline algorithms:

\begin{itemize}
    \item Empirical Risk Minimization (\textbf{ERM}, \citet{vapnik1998statistical}) minimizes the sum of errors across domains and examples.
    \item Group Distributionally Robust Optimization (\textbf{DRO}, \citet{sagawa2019distributionally}) performs ERM while increasing the importance of domains with larger errors. 
    \item Inter-domain Mixup (\textbf{Mixup}, \citet{xu2019adversarial, yan2020improve, wang2020h}) performs ERM on linear interpolations of examples from random pairs of domains and their labels.
    \item Meta-Learning for Domain Generalization (\textbf{MLDG}, \citet{li2018learning}) leverages MAML \citep{finn2017model} to meta-learn how to generalize across domains. 
    \item Different variants of the popular algorithm of \citet{ganin2016domain} to learn features $\phi(X^d)$ with distributions matching across domains:
    \begin{itemize}
        \item Domain-Adversarial Neural Networks (\textbf{DANN}, \citet{ganin2016domain}) employ  an adversarial network to match feature distributions.
        \item Class-conditional DANN (\textbf{C-DANN}, \citet{li2018deep}) is a variant of DANN matching the conditional distributions $P(\phi(X^d) | Y^d = y)$ across domains, for all labels $y$.
        \item \textbf{CORAL} \citep{sun2016deep} matches the mean and covariance of feature distributions. 
        \item \textbf{MMD} \citep{li2018domain} matches the MMD \citep{gretton2012kernel} of feature distributions.
    \end{itemize}
    \item Invariant Risk Minimization (\textbf{IRM} \citep{arjovsky2019invariant}) learns a feature representation $\phi(X^d)$ such that the optimal linear classifier on top of that representation matches across domains.
\end{itemize}

Appendix~\ref{sec:architectures} describes the network architectures and hyperparameter search spaces for all algorithms.

\subsection{Implementation choices for realistic evaluation}

Our goal is a realistic evaluation of domain generalization algorithms.
To that end, we make several implementation choices which depart from prior work, explained below.

\paragraph{Large models}
Most prior work on VLCS and PACS borrows features from or finetune ResNet-18 models \citep{He_2016_CVPR}.
Since larger ResNets are known to generalize better, we opt to finetune ResNet-50 models for all datasets except Rotated MNIST and Colored MNIST, where we use a smaller CNN architecture (see Appendix \ref{sec:architectures}).

\paragraph{Data augmentation}
Data augmentation is a standard ingredient to train image classification models.
In domain generalization, data augmentation can play an especially important role when augmentations can approximate some of the variations between domains.
Therefore, for all non-MNIST datasets, we train using the following data augmentations:
crops of random size and aspect ratio, resizing to $224 \times 224$ pixels, random horizontal flips, random color jitter, grayscaling the image with 10\% probability, and normalization using the ImageNet channel means and standard deviations.
For MNIST datasets, we use no data augmentation.

\paragraph{Using all available data}
In Rotated MNIST, whereas the usual version of the dataset constructs all domains from the same set of 1000 digits, we divide all the MNIST digits evenly among domains.
We deviate from standard practice for two reasons: we believe that using the same digits across training and test domains amounts to leaking test data, and we believe that artificially restricting the available training domain data complicates the task in an unrealistic way.

\section{Experiments}
\label{sec:experiments}

We run experiments for all algorithms (Section~\ref{sec:algorithms}), datasets (Section~\ref{sec:datasets}), and model selection criteria (Section~\ref{sec:hparam_selection}) shipped in \domainbed.
We consider all configurations of a dataset where we hide one domain for testing and train on the remaining ones.

\paragraph{Hyperparameter search}
For each algorithm and test environment, we conduct a random search \citep{bergstra2012random} of 20 trials over the hyperparameter distribution (see Appendix~\ref{sec:architectures}).
We use each model selection method from Section \ref{sec:hparam_selection} to select amongst the 20 models from the random search.
We split the data from each domain into 80\% and 20\% splits.
We use the larger splits for training and final evaluation, and the smaller splits to select hyperparameters.

\paragraph{Standard error bars}
While some domain generalization literature reports error bars across seeds, randomness arising from model selection is often ignored.
While this is acceptable if the goal is a best-versus-best comparison, it prohibits nuanced analyses.
For instance, does method A outperform method B only because random search for A got lucky?
We therefore repeat \textit{our entire study} three times making every random choice anew: hyperparameters, weight initializations, and dataset splits.
Every number we report is a mean over these repetitions, together with their estimated standard error.

This experimental protocol amounts to training a total of 45,900 neural networks.

\subsection{Results}

\begin{table}
\caption{Average out-of-distribution test accuracies for all algorithms, datasets and model selection criteria included in the initial release of \domainbed. 
These experiments compare nine popular domain generalization algorithms in the exact same conditions, showing the state-of-the-art performance of ERM.    
For a comparison against the numbers reported for thirty other algorithms in the previous literature, we refer the reader to Appendix~\ref{app:sota_review}.}
\label{table:main_results}
\rowcolors{2}{alternateRowColor}{white}
\begin{center}
\resizebox{\textwidth}{!}{%
\begin{tabular}{lcccccccc}
\toprule
\rowcolor{white}
\multicolumn{8}{c}{{Model selection method: training domain validation set}} \\
\midrule
\textbf{Algorithm}    & \textbf{CMNIST}       & \textbf{RMNIST}     & \textbf{VLCS}             & \textbf{PACS}             & \textbf{Office-Home}       & \textbf{TerraInc}   & \textbf{DomainNet} & \textbf{Avg} \\
\coloredMidrule{white}{alternateRowColor}
ERM                       & 52.0 $\pm$ 0.1            & 98.0 $\pm$ 0.0            & 77.4 $\pm$ 0.3            & 85.7 $\pm$ 0.5            & 67.5 $\pm$ 0.5            & 47.2 $\pm$ 0.4            & 41.2 $\pm$ 0.2 & 67.0\\
IRM                       & 51.8 $\pm$ 0.1            & 97.9 $\pm$ 0.0            & 78.1 $\pm$ 0.0            & 84.4 $\pm$ 1.1            & 66.6 $\pm$ 1.0            & 47.9 $\pm$ 0.7            & 35.7 $\pm$ 1.9 & 66.0\\
DRO                       & 52.0 $\pm$ 0.1            & 98.1 $\pm$ 0.0            & 77.2 $\pm$ 0.6            & 84.1 $\pm$ 0.4            & 66.9 $\pm$ 0.3            & 47.0 $\pm$ 0.3            & 33.7 $\pm$ 0.2 & 65.5\\
Mixup                     & 51.9 $\pm$ 0.1            & 98.1 $\pm$ 0.0            & 77.7 $\pm$ 0.4            & 84.3 $\pm$ 0.5            & 69.0 $\pm$ 0.1            & 48.9 $\pm$ 0.8            & 39.6 $\pm$ 0.1 & 67.1\\
MLDG                      & 51.6 $\pm$ 0.1            & 98.0 $\pm$ 0.0            & 77.1 $\pm$ 0.4            & 84.8 $\pm$ 0.6            & 68.2 $\pm$ 0.1            & 46.1 $\pm$ 0.8            & 41.8 $\pm$ 0.4 & 66.8\\
CORAL                     & 51.7 $\pm$ 0.1            & 98.1 $\pm$ 0.1            & 77.7 $\pm$ 0.5            & 86.0 $\pm$ 0.2            & 68.6 $\pm$ 0.4            & 46.4 $\pm$ 0.8            & 41.8 $\pm$ 0.2 & 67.2\\
MMD                       & 51.8 $\pm$ 0.1            & 98.1 $\pm$ 0.0            & 76.7 $\pm$ 0.9            & 85.0 $\pm$ 0.2            & 67.7 $\pm$ 0.1            & 49.3 $\pm$ 1.4            & 39.4 $\pm$ 0.8 & 66.8\\
DANN                      & 51.5 $\pm$ 0.3            & 97.9 $\pm$ 0.1            & 78.7 $\pm$ 0.3            & 84.6 $\pm$ 1.1            & 65.4 $\pm$ 0.6            & 48.4 $\pm$ 0.5            & 38.4 $\pm$ 0.0 & 66.4\\
C-DANN                    & 51.9 $\pm$ 0.1            & 98.0 $\pm$ 0.0            & 78.2 $\pm$ 0.4            & 82.8 $\pm$ 1.5            & 65.6 $\pm$ 0.5            & 47.6 $\pm$ 0.8            & 38.9 $\pm$ 0.1 & 66.1\\
\coloredBottomrule{alternateRowColor}
\end{tabular}}
\end{center}

\begin{center}
\resizebox{\textwidth}{!}{%
\begin{tabular}{lcccccccc}
\toprule
\rowcolor{white}
\multicolumn{8}{c}{{Model selection method: Leave-one-domain-out cross-validation}} \\
\midrule
\textbf{Algorithm}    & \textbf{CMNIST}       & \textbf{RMNIST}     & \textbf{VLCS}             & \textbf{PACS}             & \textbf{Office-Home}       & \textbf{TerraInc}   & \textbf{DomainNet} & \textbf{Avg} \\
\coloredMidrule{white}{alternateRowColor}
ERM                       & 34.2 $\pm$ 1.2            & 98.0 $\pm$ 0.0            & 76.8 $\pm$ 1.0            & 83.3 $\pm$ 0.6            & 67.3 $\pm$ 0.3            & 46.2 $\pm$ 0.2            & 40.8 $\pm$ 0.2 & 63.8\\
IRM                       & 36.3 $\pm$ 0.4            & 97.7 $\pm$ 0.1            & 77.2 $\pm$ 0.3            & 82.9 $\pm$ 0.6            & 66.7 $\pm$ 0.7            & 44.0 $\pm$ 0.7            & 35.3 $\pm$ 1.5 & 62.9\\
DRO                       & 32.2 $\pm$ 3.7            & 97.9 $\pm$ 0.1            & 77.5 $\pm$ 0.1            & 83.1 $\pm$ 0.6            & 67.1 $\pm$ 0.3            & 42.5 $\pm$ 0.2            & 32.8 $\pm$ 0.2 & 61.8\\
Mixup                     & 31.2 $\pm$ 2.1            & 98.1 $\pm$ 0.1            & 78.6 $\pm$ 0.2            & 83.7 $\pm$ 0.9            & 68.2 $\pm$ 0.3            & 46.1 $\pm$ 1.6            & 39.4 $\pm$ 0.3 & 63.6\\
MLDG                      & 36.9 $\pm$ 0.2            & 98.0 $\pm$ 0.1            & 77.1 $\pm$ 0.6            & 82.4 $\pm$ 0.7            & 67.6 $\pm$ 0.3            & 45.8 $\pm$ 1.2            & 42.1 $\pm$ 0.1 & 64.2\\
CORAL                     & 29.9 $\pm$ 2.5            & 98.1 $\pm$ 0.1            & 77.0 $\pm$ 0.5            & 83.6 $\pm$ 0.6            & 68.6 $\pm$ 0.2            & 48.1 $\pm$ 1.3            & 41.9 $\pm$ 0.2 & 63.9\\
MMD                       & 42.6 $\pm$ 3.0            & 98.1 $\pm$ 0.1            & 76.7 $\pm$ 0.9            & 82.8 $\pm$ 0.3            & 67.1 $\pm$ 0.5            & 46.3 $\pm$ 0.5            & 39.3 $\pm$ 0.9 & 64.7\\
DANN                      & 29.0 $\pm$ 7.7            & 89.1 $\pm$ 5.5            & 77.7 $\pm$ 0.3            & 84.0 $\pm$ 0.5            & 65.5 $\pm$ 0.1            & 45.7 $\pm$ 0.8            & 37.5 $\pm$ 0.2 & 61.2\\
C-DANN                    & 31.1 $\pm$ 8.5            & 96.3 $\pm$ 1.0            & 74.0 $\pm$ 1.0            & 81.7 $\pm$ 1.4            & 64.7 $\pm$ 0.4            & 40.6 $\pm$ 1.8            & 38.7 $\pm$ 0.2 & 61.1\\
\coloredBottomrule{alternateRowColor}
\end{tabular}}
\end{center}

\begin{center}
\resizebox{\textwidth}{!}{%
\begin{tabular}{lcccccccc}
\toprule
\rowcolor{white}
\multicolumn{8}{c}{{Model selection method: Test-domain validation set \textit{(oracle)}}} \\
\midrule
\textbf{Algorithm}    & \textbf{CMNIST}       & \textbf{RMNIST}     & \textbf{VLCS}             & \textbf{PACS}             & \textbf{Office-Home}       & \textbf{TerraInc}   & \textbf{DomainNet} & \textbf{Avg} \\
\coloredMidrule{white}{alternateRowColor}
ERM                       & 58.5 $\pm$ 0.3            & 98.1 $\pm$ 0.1            & 77.8 $\pm$ 0.3            & 87.1 $\pm$ 0.3            & 67.1 $\pm$ 0.5            & 52.7 $\pm$ 0.2            & 41.6 $\pm$ 0.1 & 68.9\\
IRM                       & 70.2 $\pm$ 0.2            & 97.9 $\pm$ 0.0            & 77.1 $\pm$ 0.2            & 84.6 $\pm$ 0.5            & 67.2 $\pm$ 0.8            & 50.9 $\pm$ 0.4            & 36.0 $\pm$ 1.6 & 69.2\\
DRO                       & 61.2 $\pm$ 0.6            & 98.1 $\pm$ 0.0            & 77.4 $\pm$ 0.6            & 87.2 $\pm$ 0.4            & 67.7 $\pm$ 0.4            & 53.1 $\pm$ 0.5            & 34.0 $\pm$ 0.1 & 68.4\\
Mixup                     & 58.4 $\pm$ 0.2            & 98.0 $\pm$ 0.0            & 78.7 $\pm$ 0.4            & 86.4 $\pm$ 0.2            & 68.5 $\pm$ 0.5            & 52.9 $\pm$ 0.3            & 40.3 $\pm$ 0.3 & 69.0\\
MLDG                      & 58.4 $\pm$ 0.2            & 98.0 $\pm$ 0.1            & 77.8 $\pm$ 0.4            & 86.8 $\pm$ 0.2            & 67.4 $\pm$ 0.2            & 52.4 $\pm$ 0.3            & 42.5 $\pm$ 0.1 & 69.1\\
CORAL                     & 57.6 $\pm$ 0.5            & 98.2 $\pm$ 0.0            & 77.8 $\pm$ 0.1            & 86.9 $\pm$ 0.2            & 68.6 $\pm$ 0.4            & 52.6 $\pm$ 0.6            & 42.1 $\pm$ 0.1 & 69.1\\
MMD                       & 63.4 $\pm$ 0.7            & 97.9 $\pm$ 0.1            & 78.0 $\pm$ 0.4            & 87.1 $\pm$ 0.5            & 67.0 $\pm$ 0.2            & 52.7 $\pm$ 0.2            & 39.8 $\pm$ 0.7 & 69.4\\
DANN                      & 58.3 $\pm$ 0.2            & 97.9 $\pm$ 0.0            & 80.1 $\pm$ 0.6            & 85.4 $\pm$ 0.7            & 65.6 $\pm$ 0.3            & 51.6 $\pm$ 0.6            & 38.3 $\pm$ 0.1 & 68.2\\
C-DANN                    & 62.0 $\pm$ 1.1            & 97.8 $\pm$ 0.1            & 80.2 $\pm$ 0.1            & 85.7 $\pm$ 0.3            & 65.6 $\pm$ 0.3            & 51.0 $\pm$ 1.0            & 38.9 $\pm$ 0.1 & 68.7\\
\coloredBottomrule{alternateRowColor}
\end{tabular}}
\end{center}
\end{table}

Table~\ref{table:main_results} summarizes the results of our experiments.
For each dataset and model, we average the best results (according to each model selection criterion) across test domains. We then report the average of \textit{this} number across three independent runs of the entire sweep, and its corresponding standard error.
For results per dataset \emph{and} domain, we refer the reader to Appendix~\ref{app:full_results}.
We draw three main conclusions from our results:

\paragraph{Our ERM baseline outperforms all previously published results}
Table \ref{table:summary} summarizes this result when model selection is performed using a training domain validation set.
What is responsible for this strong performance?
We suspect four factors: a bigger network architecture (ResNet-50), strong data augmentations, careful hyperparameter tuning and, in Rotated MNIST, using the full training data to construct our domains (instead of using a 1000-image subset).
While we are not first to use any these techniques alone, we may be first to combine all of them.
Interestingly, these results suggest standard techniques to improve in-distribution generalization are very effective at improving out-of-distribution generalization.
Our result does not refute prior work: it is possible that with similar techniques, some competing methods may improve upon ERM.
Rather, our results highlight the importance of comparing domain generalization algorithms to strong and realistic baselines.
Incorporating novel algorithms into \domainbed is an easy way to do so.
For an extensive review of results published in the literature about more than thirty algorithms, we refer the reader to Appendix~\ref{app:sota_review}.

\paragraph{When all conditions are equal, no algorithm outperforms ERM by a significant margin}
We observe this result in Table~\ref{table:main_results}, obtained from running from scratch every combination of dataset, algorithm, and model selection criteria included in \domainbed.
Given any model selection criterion, no method improves upon the average performance of ERM in more than one point.
We do not claim that any of these algorithms cannot possibly improve upon ERM, but getting substantial domain generalization improvements over ERM on these datasets proved challenging. 

\paragraph{Model selection methods matter}
We observe that model selection with a training domain validation set outperforms leave-one-domain-out cross-validation across multiple datasets and algorithms.
This does not mean that using a training domain validation set is \emph{the right way} to tune hyperparameters.
After all, it did not enable any algorithm to significantly outperform the ERM baseline.
Moreover, the stronger performance of oracle-selection (+2\%) suggests possible headroom for improvement. 

\section{Outlook}
\label{sec:outlook}

We have conducted an extensive empirical evaluation of domain generalization algorithms.
Our results led to two major conclusions.
First, empirical risk minimization achieves state-of-the-art performance when compared to eight popular domain generalization alternatives, also improving upon all the numbers previously reported in the literature.
Second, model selection has a significant effect on domain generalization, and it should be regarded as an integral part of any proposed method.
We conclude with a series of mini-discussions that answer some questions, but raise even more.

\paragraph{How can we push data augmentation further?}
While conducting our experiments, we became aware of the power of data augmentation.
\citet{zhang2019unseen} show that strong data augmentation can improve out-of-distribution generalization while not impacting in-distribution generalization.
We think of data augmentation as \emph{feature removal}: the more we augment a training example, the more invariant we make our predictor with respect to the applied transformations.
If the practitioner is lucky and performs the data augmentations that cancel the spurious correlations varying from domain to domain, then out-of-distribution performance should improve.
Given a particular domain generalization problem, what sort of data augmentation pipelines should we implement?

\paragraph{Is this as good as it gets?}
We question whether domain generalization is expected in the considered datasets.
Why do we assume a neural network should be able to classify cartoons, given only photorealistic training data?
In the case of Rotated MNIST, do truly rotation-invariant features discriminative of the digit class exist? Are those features expressible by a neural network?
Even in the presence of correct model selection, is the out-of-distribution performance of modern ERM implementations as good as it gets?
Or is it simply as bad as every other alternative?
How can we establish upper-bounds on what performance is achievable out-of-distribution via domain generalization techniques?

\paragraph{Are these the right datasets?}
Some of the datasets considered in the domain-generalization literature do not reflect realistic situations.
In reality, if one wanted to classify cartoons, the easiest option would be to collect a small labeled dataset of cartoons.
Should we consider more realistic, impactful tasks for better research in domain generalization? 
Attractive alternatives include medical imaging in different hospitals and self-driving cars in different cities.

\paragraph{It is all about (untestable) assumptions}
Every time we use ERM, we assume that training and testing examples are drawn from the same distribution.
Also every time, this is an untestable assumption.
The same applies for domain generalization: each algorithm assumes a different (untestable) type of invariance across domains.
Therefore, the performance of a domain generalization algorithm depends on the problem at hand, and only time can tell if we have made a good choice.
This is akin to the generalization of a scientific theory such as Newton's gravitation, which cannot be proved but has so far resisted falsification.
We believe there is promise in algorithms with self-adaptation capabilities during test time.

\paragraph{Benchmarking and the rules of the game}
While limiting the use of modern techniques cheapens experiments, it also distorts them from more realistic scenarios, which is the focus of our study.
Our view is that benchmark designers should balance these factors to promote a set of \emph{rules of the game} that are not only well-defined, but realistic and well-motivated.
Synthetic datasets are helpful tools, but we must not lose sight of the goal, which is artificial intelligence able to generalize in the real world.
In words of Marcel Proust:

\begin{center}
\textit{Perhaps the immobility of the things that surround us is forced upon them by our conviction that they are themselves, and not anything else, and by the immobility of our conceptions of them.}
\end{center}

\section*{Broader impact}

Current machine learning systems fail capriciously when facing novel distributions of examples.
This unreliability hinders the application of machine learning systems in critical applications such as transportation, security, and healthcare.
Here we strive to find robust machine learning models that discard spurious correlations, as we expect invariant patterns to generalize out-of-distribution. 
This should lead to fairer, safer, and more reliable machine learning systems.
But with great powers comes great responsibility: researchers in domain generalization must adhere to the strictest standards of model selection and evaluation.
We hope that our results and the release of \domainbed are some small steps in this direction, and we look forward to collaborate with fellow researchers to streamline reproducible and rigorous research towards true generalization power.

\bibliographystyle{plainnat}
\bibliography{paper}

\clearpage
\newpage
\appendix

\section{A decade of literature on domain generalization}

In this section, we provide an exhaustive literature review on a decade of domain generalization research.
The following classifies domain generalization algorithms according into four strategies to learn invariant predictors: learning invariant features, sharing parameters, meta-learning, or performing data augmentation.

\subsection{Learning invariant features}

\citet{muandet2013domain} use kernel methods to find a feature transformation that (i) minimizes the distance between transformed feature distributions across domains, and (ii) does not destroy any of the information between the original features and the targets.
In their pioneering work, \citet{ganin2016domain} propose Domain Adversarial Neural Networks (DANN), a domain adaptation technique which uses generative adversarial networks (GANs, \citet{goodfellow2014generative}), to learn a feature representation that matches across training domains.
\citet{akuzawa2019adversarial} extend DANN by considering cases where there exists an statistical dependence between the domain and the class label variables.
\citet{albuquerque2019a} extend DANN by considering one-versus-all adversaries that try to predict to which training domain does each of the examples belong to.
\citet{li2018domain} employ GANs and the maximum mean discrepancy criteria \citep{gretton2012kernel} to align feature distributions across domains.
\citet{matsuura2019domain} leverages clustering techniques to learn domain-invariant features even when the separation between training domains is not given.
\citet{li2018domain2, li2018deep} learns a feature transformation $\phi$ such that the conditional distributions $P(\phi(X^d) \mid Y^d=y)$ match for all training domains $d$ and label values $y$.
\citet{shankar2018generalizing} use a domain classifier to construct adversarial examples for a label classifier, and use a label classifier to construct adversarial examples for the domain classifier. This results in a label classifier with better domain generalization. 
\citet{li2019episodic} train a robust feature extractor and classifier. The robustness comes from (i) asking the feature extractor to produce features such that a classifier trained on domain $d$ can classify instances for domain $d' \neq d$, and (ii) asking the classifier to predict labels on domain $d$ using features produced by a feature extractor trained on domain $d' \neq d$. 
\citet{li2020sequential} adopt a lifelong learning strategy to attack the problem of domain generalization.
\citet{motiian2017unified} learn a feature representation such that (i) examples from different domains but the same class are close, (ii) examples from different domains and classes are far, and (iii) training examples can be correctly classified.
\citet{ilse2019diva} train a variational autoencoder \citep{kingma2013auto} where the bottleneck representation factorizes knowledge about domain, class label, and residual variations in the input space.
\citet{fang2013unbiased} learn a structural SVM metric such that the neighborhood of each example contains examples from the same category and all training domains.
The algorithms of \citet{sun2016deep, sun2016return, rahman2019correlation} match the feature covariance (second order statistics) across training domains at some level of representation.
The algorithms of \citet{ghifary2016scatter, hu2019domain} use kernel-based multivariate component analysis to minimize the mismatch between training domains while maximizing class separability.

Although popular, learning domain-invariant features has received some criticism \citep{zhao2019learning, johansson2019support}. Some alternatives exist, as we review next.
\citet{peters2016causal, rojas2018invariant} considered that one should search for features that lead to the same optimal classifier across training domains. In their pioneering work, \citet{peters2016causal} linked this type of invariance to the causal structure of data, and provided a basic algorithm to learn invariant linear models, based on feature selection.
\citet{arjovsky2019invariant} extend the previous to general gradient-based models, including neural networks, in their Invariant Risk Minimization (IRM) principle.
\citet{teney2020unshuffling} build on IRM to learn a feature transformation that minimizes the relative variance of classifier weights across training datasets. The authors apply their method to reduce the learning of spurious correlations in Visual Question Answering (VQA) tasks.
\citet{ahuja2020invariant} analyze IRM under a game-theoretic perspective to develop an alternative algorithm.
\citet{krueger2020out} propose an approximation to the IRM problem consisting in reducing the variance of error averages across domains.
\citet{bouvier2019hidden} attack the same problem as IRM by re-weighting data samples.

\subsection{Sharing parameters}

\citet{blanchard2011generalizing} build classifiers $f(x^d, \mu^d)$, where $\mu^d$ is a kernel mean embedding \citep{muandet2017kernel} that summarizes the dataset associated to the example $x^d$. Since the distributional identity of test instances is unknown, these embeddings are estimated using single test examples at test time. See \citet{blanchard2017domain, deshmukh2019generalization} for theoretical results on this family of algorithms.
\citet{khosla2012undoing} learn one max-margin linear classifier $w^d = w + \Delta^d$ per domain $d$, from which they distill their final, invariant predictor $w$.
\citet{ghifary2015domain} use a multitask autoencoder to learn invariances across domains. To achieve this, the authors assume that each training dataset contains the same examples; for instance, photographs about the same objects under different views.
\citet{mancini2018robust} train a deep neural network with one set of dedicated batch-normalization layers \citep{ioffe2015batch} per training dataset. Then, a softmax domain classifier predicts how to linearly-combine the batch-normalization layers at test time.
Similarly, \citet{mancini2018best} learn a softmax domain classifier used to linearly-combine domain-specific predictors at test time.
\citet{d2018domain} explore more sophisticated ways of aggregating domain-specific predictors. 
\citet{Li_2017_ICCV} extends \citet{khosla2012undoing} to deep neural networks by extending each of their parameter tensors with one additional dimension, indexed by the training domains, and set to a neutral value to predict domain-agnostic test examples.
\citet{ding2017deep} implement parameter-tying and low-rank reconstruction losses to learn a predictor that relies on common knowledge across training domains.
\citet{hu2016does, sagawa2019distributionally} weight the importance of the minibatches of the training distributions proportional to their error.

\subsection{Meta-learning}

\citet{li2018learning} employ Model-Agnostic Meta-Learning, or MAML \citep{finn2017model}, to build a predictor that learns how to adapt fast between training domains.
\citet{dou2019domain} use a similar MAML strategy, together with two regularizers that encourage  features from different domains to respect inter-class relationships, and be compactly clustered by class labels.
\citet{li2019feature} extend the MAML meta-learning strategy to instances of domain generalization where the categories vary from domain to domain.
\citet{balaji2018metareg} use MAML to meta-learn a regularizer encouraging the model trained on one domain to perform well on another domain.

\subsection{Augmenting data}

Data augmentation is an effective strategy to address domain generalization \citep{zhang2019unseen}. Unfortunately, how to design efficient data augmentation routines depends on the type of data at hand, and demands a significant amount of work from human experts.
\citet{xu2019adversarial, yan2020improve, wang2020h} use mixup \citep{zhang2017mixup} to blend examples from the different training distributions.
\citet{carlucci2019domain} constructs an auxiliary classification task aimed at solving jigsaw puzzles of image patches. The authors show that this self-supervised learning task learns features that improve domain generalization.
\citet{albuquerque2020i} introduce the self-supervised task of predicting responses to Gabor filter banks, in order to learn more transferrable features.
\citet{wang2019learning} remove textural information from images to improve domain generalization.
\citet{volpi2018generalizing} show that training with adversarial data augmentation on a single domain is sufficient to improve domain generalization.
\citet{nam2019reducing, asadi2019towards} promote representations of data that ignore texture and focus on shape.
\citet{rahman2019multi, zhou2020deep, carlucci2019domain} are three alternatives that use GANs to augment the data available during training time.

\subsection{Previous state-of-the-art numbers}
\label{app:sota_review}

Table~\ref{table:sota_review} compiles the best out-of-distribution test accuracies reported across a decade of domain generalization research.

\begin{table}
\caption{Previous state-of-the-art in the literature of domain generalization.}
\begin{center}
\begin{small}
\resizebox{\textwidth}{!}{
\begin{tabular}{p{1.5cm}cccccccl}
\toprule
\textbf{Benchmark} & \multicolumn{7}{c}{\textbf{Accuracy (by domain)}} & \textbf{Algorithm}\\
\midrule
\multirow{11}{1.5cm}{Rotated MNIST}%
&     0 &    15 &    30 &    45 &    60 &    75 & Average & \\
& 82.50 & 96.30 & 93.40 & 78.60 & 94.20 & 80.50 & 87.58 & D-MTAE \citep{ghifary2015domain}\\
& 84.60 & 95.60 & 94.60 & 82.90 & 94.80 & 82.10 & 89.10 & CCSA \citep{motiian2017unified}\\
& 83.70 & 96.90 & 95.70 & 85.20 & 95.90 & 81.20 & 89.80 & MMD-AAE \citep{li2018domain}\\
& 85.60 & 95.00 & 95.60 & 95.50 & 95.90 & 84.30 & 92.00 & BestSources \citep{mancini2018best}\\
& 88.80 & 97.60 & 97.50 & 97.80 & 97.60 & 91.90 & 95.20 & ADAGE \citep{carlucci2019h}\\
& 88.30 & 98.60 & 98.00 & 97.70 & 97.70 & 91.40 & 95.28 & CrossGrad \citep{shankar2018generalizing}\\
& 90.10 & 98.90 & 98.90 & 98.80 & 98.30 & 90.00 & 95.80 & HEX \citep{wang2019learning}\\
& 89.23 & 99.68 & 99.20 & 99.24 & 99.53 & 91.44 & 96.39 & FeatureCritic \citep{li2019feature}\\
& 93.50 & 99.30 & 99.10 & 99.20 & 99.30 & 93.00 & 97.20 & DIVA \citep{ilse2019diva}\\
\midrule
\multirow{18}{*}{VLCS}%
&     C &     L &     S &     V &       &       & Average & \\
& 88.92 & 59.60 & 59.20 & 64.36 &       &       & 64.06 & SCA \citep{ghifary2016scatter}\\
& 92.30 & 62.10 & 59.10 & 67.10 &       &       & 65.00 & CCSA \citep{motiian2017unified}\\
& 89.15 & 64.99 & 58.88 & 62.59 &       &       & 67.67 & MTSSL \citep{albuquerque2020i}\\
& 89.05 & 60.13 & 61.33 & 63.90 &       &       & 68.60 & D-MTAE \citep{ghifary2015domain}\\
& 91.12 & 60.43 & 60.85 & 65.65 &       &       & 69.41 & CIDG \citep{li2018domain2}\\
& 88.83 & 63.06 & 62.10 & 64.38 &       &       & 69.59 & CIDDG \citep{li2018deep}\\
& 92.64 & 61.78 & 59.60 & 66.86 &       &       & 70.22 & MDA \citep{hu2019domain}\\
& 92.76 & 62.34 & 63.54 & 65.25 &       &       & 70.97 & MDA \citep{ding2017deep}\\
& 93.63 & 63.49 & 61.32 & 69.99 &       &       & 72.11 & DBADG \citep{Li_2017_ICCV}\\
& 94.40 & 62.60 & 64.40 & 67.60 &       &       & 72.30 & MMD-AAE \citep{li2018domain}\\
& 94.10 & 64.30 & 65.90 & 67.10 &       &       & 72.90 & Epi-FCR \citep{li2019episodic}\\
& 96.93 & 60.90 & 64.30 & 70.62 &       &       & 73.19 & JiGen \citep{carlucci2019domain}\\
& 96.72 & 60.40 & 63.68 & 70.49 &       &       & 73.30 & REx \citep{krueger2020out}\\
& 96.40 & 64.80 & 64.00 & 68.70 &       &       & 73.50 & S-MLDG \citep{li2020sequential}\\
& 96.66 & 58.77 & 68.13 & 71.96 &       &       & 73.88 & MMLD \citep{matsuura2019domain}\\
& 94.78 & 64.90 & 67.64 & 69.14 &       &       & 74.11 & MASF \citep{dou2019domain}\\
& 98.11 & 63.61 & 67.11 & 74.33 &       &       & 75.79 & DDEC \citep{asadi2019towards}\\
& 95.52 & 67.63 & 69.37 & 71.14 &       &       & 75.92 & ATIR \citep{albuquerque2019a}\\
\midrule
\multirow{20}{*}{PACS}%
&     A &     C &     P &     S &       &       & Average & \\
& 62.86 & 66.97 & 89.50 & 57.51 &       &       & 69.21 & DBADG \citep{Li_2017_ICCV}\\
& 61.67 & 67.41 & 84.31 & 63.91 &       &       & 69.32 & MTSSL \citep{albuquerque2020i}\\
& 62.70 & 69.73 & 78.65 & 64.45 &       &       & 69.40 & CIDDG \citep{li2018deep}\\
& 62.64 & 65.98 & 90.44 & 58.76 &       &       & 69.45 & JAN-COMBO \citep{rahman2019multi}\\
& 66.23 & 66.88 & 88.00 & 58.96 &       &       & 70.01 & MLDG \citep{li2018learning}\\
& 66.80 & 69.70 & 87.90 & 56.30 &       &       & 70.20 & HEX \citep{wang2019learning}\\
& 64.10 & 66.80 & 90.20 & 60.10 &       &       & 70.30 & BestSoruces \citep{mancini2018best}\\
& 64.40 & 68.60 & 90.10 & 58.40 &       &       & 70.40 & FeatureCritic \citep{li2019feature}\\
& 67.04 & 67.97 & 89.74 & 59.81 &       &       & 71.14 & REx \citep{krueger2020out}\\
& 65.52 & 69.90 & 89.16 & 63.37 &       &       & 71.98 & CAADG \citep{rahman2019correlation}\\
& 64.70 & 72.30 & 86.10 & 65.00 &       &       & 72.00 & Epi-FCR \citep{li2019episodic}\\
& 66.60 & 73.36 & 88.12 & 66.19 &       &       & 73.55 & ATIR \citep{albuquerque2019a}\\
& 70.35 & 72.46 & 90.68 & 67.33 &       &       & 75.21 & MASF \citep{dou2019domain}\\
& 79.42 & 75.25 & 96.03 & 71.35 &       &       & 80.51 & JiGen \citep{carlucci2019domain}\\
& 80.50 & 77.80 & 94.80 & 72.80 &       &       & 81.50 & S-MLDG \citep{li2020sequential}\\
& 79.48 & 77.13 & 94.30 & 75.30 &       &       & 81.55 & D-SAM-$\Lambda$ \citep{d2018domain}\\
& 84.20 & 78.10 & 95.30 & 74.70 &       &       & 83.10 & DDAIG \citep{zhou2020deep}\\
& 81.28 & 77.16 & 96.09 & 72.29 &       &       & 81.83 & MMLD \citep{matsuura2019domain}\\
& 83.58 & 77.66 & 95.47 & 76.30 &       &       & 83.25 & SagNets \citep{nam2019reducing}\\
& 87.20 & 79.20 & 97.60 & 70.30 &       &       & 83.60 & MetaReg \citep{balaji2018metareg}\\
& 83.01 & 79.39 & 96.83 & 78.62 &       &       & 84.46 & DDEC \citep{asadi2019towards}\\
\midrule
\multirow{5}{1.5cm}{Office Home}%
&     A &     C &     P &     R &       &       & Average & \\
& 48.09 & 45.20 & 66.52 & 68.35 &       &       & 57.04 & JAN-COMBO \citep{rahman2019multi}\\
& 53.04 & 47.51 & 71.47 & 72.79 &       &       & 61.20 & JiGen \citep{carlucci2019domain}\\
& 54.53 & 49.04 & 71.57 & 71.90 &       &       & 61.76 & D-SAM-$\Lambda$ \citep{d2018domain}\\
& 60.20 & 45.38 & 70.42 & 73.38 &       &       & 62.34 & SagNets \citep{nam2019reducing}\\
& 59.20 & 52.30 & 74.60 & 76.00 &       &       & 65.50 & DDAIG \citep{zhou2020deep}\\
\bottomrule
\end{tabular}
}
\end{small}
\end{center}
\label{table:sota_review}
\end{table}

\clearpage
\newpage
\section{Domain generalization accuracies per algorithm, dataset, and domain}
\label{app:full_results}
\subsection{Colored MNIST}
\begin{center}
\begin{tabular}{lccc}
\toprule
\multicolumn{4}{c}{\textbf{Model selection method: training domain validation set}} \\
\midrule
\textbf{Algorithm}    & \textbf{0.1}              & \textbf{0.2}              & \textbf{0.9}              \\
\midrule
ERM                       & 72.7 $\pm$ 0.2            & 73.2 $\pm$ 0.3            & 10.0 $\pm$ 0.0            \\
IRM                       & 72.0 $\pm$ 0.3            & 73.2 $\pm$ 0.0            & 10.1 $\pm$ 0.2            \\
DRO                 & 72.7 $\pm$ 0.3            & 73.1 $\pm$ 0.3            & 10.0 $\pm$ 0.0            \\
Mixup                     & 72.4 $\pm$ 0.2            & 73.3 $\pm$ 0.3            & 10.0 $\pm$ 0.1            \\
MLDG                      & 71.4 $\pm$ 0.4            & 73.3 $\pm$ 0.0            & 10.0 $\pm$ 0.0            \\
CORAL                     & 71.8 $\pm$ 0.4            & 73.3 $\pm$ 0.2            & 10.1 $\pm$ 0.1            \\
MMD                       & 72.1 $\pm$ 0.2            & 72.8 $\pm$ 0.2            & 10.5 $\pm$ 0.2            \\
ADA                       & 72.0 $\pm$ 0.3            & 72.4 $\pm$ 0.5            & 10.0 $\pm$ 0.2            \\
CondADA                   & 72.2 $\pm$ 0.3            & 73.2 $\pm$ 0.2            & 10.4 $\pm$ 0.3            \\
\bottomrule
\end{tabular}
\end{center}

\begin{center}
\begin{tabular}{lccc}
\toprule
\multicolumn{4}{c}{\textbf{Model selection method: leave-one-domain-out cross-validation}} \\
\midrule
\textbf{Algorithm}    & \textbf{0.1}              & \textbf{0.2}              & \textbf{0.9}              \\
\midrule
ERM                       & 46.0 $\pm$ 3.4            & 46.6 $\pm$ 3.8            & 10.0 $\pm$ 0.1            \\
IRM                       & 49.3 $\pm$ 0.9            & 49.5 $\pm$ 0.2            & 10.0 $\pm$ 0.2            \\
DRO                 & 36.7 $\pm$ 10.9           & 49.9 $\pm$ 0.3            & 10.2 $\pm$ 0.1            \\
Mixup                     & 43.0 $\pm$ 5.1            & 40.7 $\pm$ 8.0            & 10.0 $\pm$ 0.1            \\
MLDG                      & 50.3 $\pm$ 0.4            & 50.1 $\pm$ 0.3            & 10.1 $\pm$ 0.1            \\
CORAL                     & 24.5 $\pm$ 10.6           & 47.2 $\pm$ 2.4            & 18.1 $\pm$ 6.8            \\
MMD                       & 50.1 $\pm$ 0.3            & 55.0 $\pm$ 4.2            & 22.8 $\pm$ 10.5           \\
DANN                       & 50.2 $\pm$ 0.3            & 39.9 $\pm$ 7.7            & 10.2 $\pm$ 0.0            \\
C-DANN                   & 50.5 $\pm$ 0.0            & 49.4 $\pm$ 0.0            & 11.8 $\pm$ 1.1            \\
\bottomrule
\end{tabular}
\end{center}

\begin{center}
\begin{tabular}{lccc}
\toprule
\multicolumn{4}{c}{\textbf{Model selection method: test-domain validation set \textit{(oracle)}}} \\
\midrule
\textbf{Algorithm}    & \textbf{0.1}              & \textbf{0.2}              & \textbf{0.9}              \\
\midrule
ERM                       & 72.3 $\pm$ 0.6            & 73.1 $\pm$ 0.3            & 30.0 $\pm$ 0.3            \\
IRM                       & 72.7 $\pm$ 0.1            & 72.8 $\pm$ 0.3            & 65.2 $\pm$ 0.8            \\
DRO                 & 73.3 $\pm$ 0.2            & 72.9 $\pm$ 0.2            & 37.4 $\pm$ 1.9            \\
Mixup                     & 72.6 $\pm$ 0.6            & 73.7 $\pm$ 0.2            & 28.8 $\pm$ 0.5            \\
MLDG                      & 71.2 $\pm$ 0.2            & 73.6 $\pm$ 0.2            & 30.4 $\pm$ 0.5            \\
CORAL                     & 71.2 $\pm$ 0.1            & 72.2 $\pm$ 0.5            & 29.4 $\pm$ 1.3            \\
MMD                       & 67.3 $\pm$ 2.6            & 72.8 $\pm$ 0.3            & 50.2 $\pm$ 0.3            \\
DANN                       & 72.6 $\pm$ 0.4            & 73.0 $\pm$ 0.3            & 29.3 $\pm$ 1.2            \\
C-DANN                   & 72.4 $\pm$ 0.0            & 73.1 $\pm$ 0.5            & 40.5 $\pm$ 3.8            \\
\bottomrule
\end{tabular}
\end{center}

\clearpage
\newpage
\subsection{Rotated MNIST}
\begin{center}
\begin{tabular}{lcccccc}
\toprule
\multicolumn{7}{c}{\textbf{Model selection method: training domain validation set}} \\
\midrule
\textbf{Algorithm}    & \textbf{0$^{\circ}$}      & \textbf{15$^{\circ}$}     & \textbf{30$^{\circ}$}     & \textbf{45$^{\circ}$}     & \textbf{60$^{\circ}$}     & \textbf{75$^{\circ}$}     \\
\midrule
ERM                       & 95.6 $\pm$ 0.1            & 99.0 $\pm$ 0.1            & 98.9 $\pm$ 0.0            & 99.1 $\pm$ 0.1            & 99.0 $\pm$ 0.0            & 96.7 $\pm$ 0.2            \\
IRM                       & 95.9 $\pm$ 0.2            & 98.9 $\pm$ 0.0            & 99.0 $\pm$ 0.0            & 98.8 $\pm$ 0.1            & 98.9 $\pm$ 0.1            & 95.5 $\pm$ 0.3            \\
DRO                 & 95.9 $\pm$ 0.1            & 98.9 $\pm$ 0.0            & 99.0 $\pm$ 0.1            & 99.0 $\pm$ 0.0            & 99.0 $\pm$ 0.0            & 96.9 $\pm$ 0.1            \\
Mixup                     & 96.1 $\pm$ 0.2            & 99.1 $\pm$ 0.0            & 98.9 $\pm$ 0.0            & 99.0 $\pm$ 0.0            & 99.0 $\pm$ 0.1            & 96.6 $\pm$ 0.1            \\
MLDG                      & 95.9 $\pm$ 0.2            & 98.9 $\pm$ 0.1            & 99.0 $\pm$ 0.0            & 99.1 $\pm$ 0.0            & 99.0 $\pm$ 0.0            & 96.0 $\pm$ 0.2            \\
CORAL                     & 95.7 $\pm$ 0.2            & 99.0 $\pm$ 0.0            & 99.1 $\pm$ 0.1            & 99.1 $\pm$ 0.0            & 99.0 $\pm$ 0.0            & 96.7 $\pm$ 0.2            \\
MMD                       & 96.6 $\pm$ 0.1            & 98.9 $\pm$ 0.0            & 98.9 $\pm$ 0.1            & 99.1 $\pm$ 0.1            & 99.0 $\pm$ 0.0            & 96.2 $\pm$ 0.1            \\
DANN                       & 95.6 $\pm$ 0.3            & 98.9 $\pm$ 0.0            & 98.9 $\pm$ 0.0            & 99.0 $\pm$ 0.1            & 98.9 $\pm$ 0.0            & 95.9 $\pm$ 0.5            \\
C-DANN                   & 96.0 $\pm$ 0.5            & 98.8 $\pm$ 0.0            & 99.0 $\pm$ 0.1            & 99.1 $\pm$ 0.0            & 98.9 $\pm$ 0.1            & 96.5 $\pm$ 0.3            \\
\bottomrule
\end{tabular}
\end{center}

\begin{center}
\begin{tabular}{lcccccc}
\toprule
\multicolumn{7}{c}{\textbf{Model selection method: leave-one-domain-out cross-validation}} \\
\midrule
\textbf{Algorithm}    & \textbf{0$^{\circ}$}      & \textbf{15$^{\circ}$}     & \textbf{30$^{\circ}$}     & \textbf{45$^{\circ}$}     & \textbf{60$^{\circ}$}     & \textbf{75$^{\circ}$}     \\
\midrule
ERM                       & 95.9 $\pm$ 0.2            & 99.0 $\pm$ 0.1            & 99.0 $\pm$ 0.0            & 99.0 $\pm$ 0.1            & 99.0 $\pm$ 0.0            & 96.3 $\pm$ 0.1            \\
IRM                       & 95.5 $\pm$ 0.4            & 98.7 $\pm$ 0.2            & 98.7 $\pm$ 0.1            & 98.5 $\pm$ 0.3            & 98.7 $\pm$ 0.1            & 96.1 $\pm$ 0.1            \\
DRO                 & 95.5 $\pm$ 0.5            & 98.4 $\pm$ 0.5            & 99.0 $\pm$ 0.1            & 99.0 $\pm$ 0.0            & 98.8 $\pm$ 0.2            & 96.6 $\pm$ 0.1            \\
Mixup                     & 95.9 $\pm$ 0.3            & 98.8 $\pm$ 0.1            & 99.0 $\pm$ 0.0            & 99.0 $\pm$ 0.0            & 99.0 $\pm$ 0.0            & 96.5 $\pm$ 0.0            \\
MLDG                      & 95.8 $\pm$ 0.4            & 98.9 $\pm$ 0.1            & 99.0 $\pm$ 0.1            & 99.0 $\pm$ 0.0            & 98.9 $\pm$ 0.0            & 96.2 $\pm$ 0.1            \\
CORAL                     & 96.2 $\pm$ 0.1            & 98.9 $\pm$ 0.1            & 99.1 $\pm$ 0.0            & 99.0 $\pm$ 0.1            & 98.7 $\pm$ 0.2            & 96.5 $\pm$ 0.2            \\
MMD                       & 96.5 $\pm$ 0.2            & 98.9 $\pm$ 0.0            & 98.8 $\pm$ 0.2            & 99.0 $\pm$ 0.1            & 98.7 $\pm$ 0.1            & 96.4 $\pm$ 0.1            \\
DANN                       & 85.5 $\pm$ 4.7            & 78.1 $\pm$ 16.5           & 98.1 $\pm$ 0.6            & 98.7 $\pm$ 0.0            & 93.8 $\pm$ 1.8            & 95.9 $\pm$ 0.7            \\
C-DANN                   & 73.7 $\pm$ 0.0            & 98.7 $\pm$ 0.0            & 98.7 $\pm$ 0.1            & 97.0 $\pm$ 0.0            & 98.3 $\pm$ 0.4            & 94.6 $\pm$ 1.2            \\
\bottomrule
\end{tabular}
\end{center}

\begin{center}
\begin{tabular}{lcccccc}
\toprule
\multicolumn{7}{c}{\textbf{Model selection method: test-domain validation set \textit{(oracle)}}} \\
\midrule
\textbf{Algorithm}    & \textbf{0$^{\circ}$}      & \textbf{15$^{\circ}$}     & \textbf{30$^{\circ}$}     & \textbf{45$^{\circ}$}     & \textbf{60$^{\circ}$}     & \textbf{75$^{\circ}$}     \\
\midrule
ERM                       & 96.0 $\pm$ 0.2            & 98.8 $\pm$ 0.1            & 98.8 $\pm$ 0.1            & 99.0 $\pm$ 0.0            & 99.0 $\pm$ 0.0            & 96.8 $\pm$ 0.1            \\
IRM                       & 96.0 $\pm$ 0.2            & 98.9 $\pm$ 0.0            & 99.0 $\pm$ 0.0            & 98.8 $\pm$ 0.1            & 98.9 $\pm$ 0.1            & 95.7 $\pm$ 0.3            \\
DRO                 & 96.2 $\pm$ 0.1            & 98.9 $\pm$ 0.0            & 99.0 $\pm$ 0.1            & 98.7 $\pm$ 0.1            & 99.1 $\pm$ 0.0            & 96.8 $\pm$ 0.1            \\
Mixup                     & 95.8 $\pm$ 0.3            & 98.9 $\pm$ 0.1            & 99.0 $\pm$ 0.1            & 99.0 $\pm$ 0.1            & 98.9 $\pm$ 0.1            & 96.5 $\pm$ 0.1            \\
MLDG                      & 96.2 $\pm$ 0.1            & 99.0 $\pm$ 0.0            & 99.0 $\pm$ 0.1            & 98.9 $\pm$ 0.1            & 99.0 $\pm$ 0.1            & 96.1 $\pm$ 0.2            \\
CORAL                     & 96.4 $\pm$ 0.1            & 99.0 $\pm$ 0.0            & 99.0 $\pm$ 0.1            & 99.0 $\pm$ 0.0            & 98.9 $\pm$ 0.1            & 96.8 $\pm$ 0.2            \\
MMD                       & 95.7 $\pm$ 0.4            & 98.8 $\pm$ 0.0            & 98.9 $\pm$ 0.1            & 98.8 $\pm$ 0.1            & 99.0 $\pm$ 0.0            & 96.3 $\pm$ 0.2            \\
DANN                       & 96.0 $\pm$ 0.1            & 98.8 $\pm$ 0.1            & 98.6 $\pm$ 0.1            & 98.7 $\pm$ 0.1            & 98.8 $\pm$ 0.1            & 96.4 $\pm$ 0.1            \\
C-DANN                   & 95.8 $\pm$ 0.2            & 98.8 $\pm$ 0.0            & 98.9 $\pm$ 0.0            & 98.6 $\pm$ 0.1            & 98.8 $\pm$ 0.1            & 96.1 $\pm$ 0.2            \\
\bottomrule
\end{tabular}
\end{center}

\clearpage
\newpage
\subsection{VLCS}
\begin{center}
\begin{tabular}{lcccc}
\toprule
\multicolumn{5}{c}{\textbf{Model selection method: training domain validation set}} \\
\midrule
\textbf{Algorithm}    & \textbf{C}                & \textbf{L}                & \textbf{S}                & \textbf{V}                \\
\midrule
ERM                       & 97.6 $\pm$ 1.0            & 63.3 $\pm$ 0.9            & 72.2 $\pm$ 0.5            & 76.4 $\pm$ 1.5            \\
IRM                       & 97.6 $\pm$ 0.3            & 65.0 $\pm$ 0.9            & 72.9 $\pm$ 0.5            & 76.9 $\pm$ 1.3            \\
DRO                 & 97.7 $\pm$ 0.4            & 62.5 $\pm$ 1.1            & 70.1 $\pm$ 0.7            & 78.4 $\pm$ 0.9            \\
Mixup                     & 97.9 $\pm$ 0.3            & 64.5 $\pm$ 0.6            & 71.5 $\pm$ 0.9            & 76.9 $\pm$ 1.3            \\
MLDG                      & 98.1 $\pm$ 0.3            & 63.0 $\pm$ 0.9            & 73.5 $\pm$ 0.6            & 73.7 $\pm$ 0.3            \\
CORAL                     & 98.8 $\pm$ 0.1            & 64.6 $\pm$ 0.8            & 71.7 $\pm$ 1.4            & 75.8 $\pm$ 0.4            \\
MMD                       & 97.1 $\pm$ 0.4            & 63.4 $\pm$ 0.7            & 71.4 $\pm$ 0.8            & 74.9 $\pm$ 2.5            \\
DANN                       & 98.5 $\pm$ 0.2            & 64.9 $\pm$ 1.1            & 73.1 $\pm$ 0.7            & 78.3 $\pm$ 0.3            \\
C-DANN                   & 97.5 $\pm$ 0.1            & 65.2 $\pm$ 0.4            & 73.4 $\pm$ 1.1            & 76.9 $\pm$ 0.2            \\
\bottomrule
\end{tabular}
\end{center}

\begin{center}
\begin{tabular}{lcccc}
\toprule
\multicolumn{5}{c}{\textbf{Model selection method: leave-one-domain-out cross-validation}} \\
\midrule
\textbf{Algorithm}    & \textbf{C}                & \textbf{L}                & \textbf{S}                & \textbf{V}                \\
\midrule
ERM                       & 97.8 $\pm$ 0.0            & 63.3 $\pm$ 1.6            & 70.3 $\pm$ 1.6            & 75.9 $\pm$ 1.4            \\
IRM                       & 98.9 $\pm$ 0.0            & 63.6 $\pm$ 0.8            & 71.1 $\pm$ 2.2            & 75.4 $\pm$ 1.5            \\
DRO                 & 99.2 $\pm$ 0.2            & 62.0 $\pm$ 1.6            & 73.4 $\pm$ 0.8            & 75.5 $\pm$ 1.0            \\
Mixup                     & 97.9 $\pm$ 0.7            & 65.5 $\pm$ 0.8            & 73.3 $\pm$ 0.8            & 77.8 $\pm$ 0.5            \\
MLDG                      & 96.3 $\pm$ 1.1            & 65.1 $\pm$ 0.9            & 71.9 $\pm$ 1.5            & 75.0 $\pm$ 0.5            \\
CORAL                     & 97.5 $\pm$ 0.1            & 64.0 $\pm$ 0.2            & 69.7 $\pm$ 2.0            & 76.7 $\pm$ 0.3            \\
MMD                       & 97.7 $\pm$ 0.4            & 63.1 $\pm$ 1.9            & 68.6 $\pm$ 1.5            & 77.5 $\pm$ 1.2            \\
DANN                       & 95.3 $\pm$ 1.8            & 61.3 $\pm$ 1.8            & 74.3 $\pm$ 1.0            & 79.7 $\pm$ 0.9            \\
C-DANN                   & 92.3 $\pm$ 4.2            & 60.3 $\pm$ 1.5            & 68.4 $\pm$ 2.1            & 74.9 $\pm$ 1.3            \\
\bottomrule
\end{tabular}
\end{center}

\begin{center}
\begin{tabular}{lcccc}
\toprule
\multicolumn{5}{c}{\textbf{Model selection method: test-domain validation set \textit{(oracle)}}} \\
\midrule
\textbf{Algorithm}    & \textbf{C}                & \textbf{L}                & \textbf{S}                & \textbf{V}                \\
\midrule
ERM                       & 97.7 $\pm$ 0.3            & 65.2 $\pm$ 0.4            & 73.2 $\pm$ 0.7            & 75.2 $\pm$ 0.4            \\
IRM                       & 97.6 $\pm$ 0.5            & 64.7 $\pm$ 1.1            & 69.7 $\pm$ 0.5            & 76.6 $\pm$ 0.7            \\
DRO                 & 97.8 $\pm$ 0.0            & 66.4 $\pm$ 0.5            & 68.7 $\pm$ 1.2            & 76.8 $\pm$ 1.0            \\
Mixup                     & 98.3 $\pm$ 0.3            & 66.7 $\pm$ 0.5            & 73.3 $\pm$ 1.1            & 76.3 $\pm$ 0.8            \\
MLDG                      & 98.4 $\pm$ 0.2            & 65.9 $\pm$ 0.5            & 70.7 $\pm$ 0.8            & 76.1 $\pm$ 0.6            \\
CORAL                     & 98.1 $\pm$ 0.1            & 67.1 $\pm$ 0.8            & 70.1 $\pm$ 0.6            & 75.8 $\pm$ 0.5            \\
MMD                       & 98.1 $\pm$ 0.3            & 66.2 $\pm$ 0.2            & 70.5 $\pm$ 1.0            & 77.2 $\pm$ 0.6            \\
DANN                       & 98.2 $\pm$ 0.3            & 67.8 $\pm$ 1.1            & 74.2 $\pm$ 0.7            & 80.1 $\pm$ 0.6            \\
C-DANN                   & 98.9 $\pm$ 0.3            & 68.8 $\pm$ 0.6            & 73.7 $\pm$ 0.6            & 79.3 $\pm$ 0.6            \\
\bottomrule
\end{tabular}
\end{center}

\clearpage
\newpage
\subsection{PACS}
\begin{center}
\begin{tabular}{lcccc}
\toprule
\multicolumn{5}{c}{\textbf{Model selection method: training domain validation set}} \\
\midrule
\textbf{Algorithm}    & \textbf{A}                & \textbf{C}                & \textbf{P}                & \textbf{S}                \\
\midrule
ERM                       & 88.1 $\pm$ 0.1            & 77.9 $\pm$ 1.3            & 97.8 $\pm$ 0.0            & 79.1 $\pm$ 0.9            \\
IRM                       & 85.0 $\pm$ 1.6            & 77.6 $\pm$ 0.9            & 96.7 $\pm$ 0.3            & 78.5 $\pm$ 2.6            \\
DRO                 & 86.4 $\pm$ 0.3            & 79.9 $\pm$ 0.8            & 98.0 $\pm$ 0.3            & 72.1 $\pm$ 0.7            \\
Mixup                     & 86.5 $\pm$ 0.4            & 76.6 $\pm$ 1.5            & 97.7 $\pm$ 0.2            & 76.5 $\pm$ 1.2            \\
MLDG                      & 89.1 $\pm$ 0.9            & 78.8 $\pm$ 0.7            & 97.0 $\pm$ 0.9            & 74.4 $\pm$ 2.0            \\
CORAL                     & 87.7 $\pm$ 0.6            & 79.2 $\pm$ 1.1            & 97.6 $\pm$ 0.0            & 79.4 $\pm$ 0.7            \\
MMD                       & 84.5 $\pm$ 0.6            & 79.7 $\pm$ 0.7            & 97.5 $\pm$ 0.4            & 78.1 $\pm$ 1.3            \\
DANN                       & 85.9 $\pm$ 0.5            & 79.9 $\pm$ 1.4            & 97.6 $\pm$ 0.2            & 75.2 $\pm$ 2.8            \\
C-DANN                   & 84.0 $\pm$ 0.9            & 78.5 $\pm$ 1.5            & 97.0 $\pm$ 0.4            & 71.8 $\pm$ 3.9            \\
\bottomrule
\end{tabular}
\end{center}

\begin{center}
\begin{tabular}{lcccc}
\toprule
\multicolumn{5}{c}{\textbf{Model selection method: leave-one-domain-out cross-validation}} \\
\midrule
\textbf{Algorithm}    & \textbf{A}                & \textbf{C}                & \textbf{P}                & \textbf{S}                \\
\midrule
ERM                       & 83.9 $\pm$ 1.6            & 78.6 $\pm$ 2.0            & 97.3 $\pm$ 0.1            & 73.5 $\pm$ 1.1            \\
IRM                       & 82.5 $\pm$ 2.6            & 78.0 $\pm$ 0.3            & 96.7 $\pm$ 1.1            & 74.4 $\pm$ 1.3            \\
DRO                 & 87.1 $\pm$ 0.3            & 77.6 $\pm$ 1.9            & 97.2 $\pm$ 0.4            & 70.7 $\pm$ 3.0            \\
Mixup                     & 88.0 $\pm$ 0.5            & 74.3 $\pm$ 4.0            & 97.2 $\pm$ 0.2            & 75.3 $\pm$ 0.2            \\
MLDG                      & 85.8 $\pm$ 0.7            & 77.3 $\pm$ 0.5            & 96.8 $\pm$ 0.5            & 69.9 $\pm$ 3.6            \\
CORAL                     & 86.0 $\pm$ 1.1            & 75.5 $\pm$ 2.3            & 96.2 $\pm$ 0.9            & 76.6 $\pm$ 2.1            \\
MMD                       & 85.9 $\pm$ 0.5            & 78.1 $\pm$ 2.1            & 96.2 $\pm$ 1.3            & 71.1 $\pm$ 3.0            \\
DANN                       & 86.7 $\pm$ 0.3            & 78.5 $\pm$ 0.5            & 97.4 $\pm$ 0.4            & 73.3 $\pm$ 2.3            \\
C-DANN                   & 83.6 $\pm$ 3.8            & 75.9 $\pm$ 1.8            & 97.4 $\pm$ 0.5            & 70.0 $\pm$ 3.6            \\
\bottomrule
\end{tabular}
\end{center}

\begin{center}
\begin{tabular}{lcccc}
\toprule
\multicolumn{5}{c}{\textbf{Model selection method: test-domain validation set \textit{(oracle)}}} \\
\midrule
\textbf{Algorithm}    & \textbf{A}                & \textbf{C}                & \textbf{P}                & \textbf{S}                \\
\midrule
ERM                       & 87.8 $\pm$ 0.4            & 82.8 $\pm$ 0.5            & 97.6 $\pm$ 0.4            & 80.4 $\pm$ 0.6            \\
IRM                       & 85.7 $\pm$ 1.0            & 79.3 $\pm$ 1.1            & 97.6 $\pm$ 0.4            & 75.9 $\pm$ 1.0            \\
DRO                 & 88.2 $\pm$ 0.7            & 82.4 $\pm$ 0.8            & 97.7 $\pm$ 0.2            & 80.6 $\pm$ 0.9            \\
Mixup                     & 87.4 $\pm$ 1.0            & 80.7 $\pm$ 1.0            & 97.9 $\pm$ 0.2            & 79.7 $\pm$ 1.0            \\
MLDG                      & 87.1 $\pm$ 0.9            & 81.3 $\pm$ 1.5            & 97.6 $\pm$ 0.4            & 81.2 $\pm$ 1.0            \\
CORAL                     & 87.4 $\pm$ 0.6            & 82.2 $\pm$ 0.3            & 97.6 $\pm$ 0.1            & 80.2 $\pm$ 0.4            \\
MMD                       & 87.6 $\pm$ 1.2            & 83.0 $\pm$ 0.4            & 97.8 $\pm$ 0.1            & 80.1 $\pm$ 1.0            \\
DANN                       & 86.4 $\pm$ 1.4            & 80.6 $\pm$ 1.0            & 97.7 $\pm$ 0.2            & 77.1 $\pm$ 1.3            \\
C-DANN                   & 87.0 $\pm$ 1.2            & 80.8 $\pm$ 0.9            & 97.4 $\pm$ 0.5            & 77.6 $\pm$ 0.1            \\
\bottomrule
\end{tabular}
\end{center}

\clearpage
\newpage
\subsection{Office-Home}
\begin{center}
\begin{tabular}{lcccc}
\toprule
\multicolumn{5}{c}{\textbf{Model selection method: training domain validation set}} \\
\midrule
\textbf{Algorithm}    & \textbf{A}                & \textbf{C}                & \textbf{P}                & \textbf{R}                \\
\midrule
ERM                       & 62.7 $\pm$ 1.1            & 53.4 $\pm$ 0.6            & 76.5 $\pm$ 0.4            & 77.3 $\pm$ 0.3            \\
IRM                       & 61.8 $\pm$ 1.0            & 52.3 $\pm$ 1.0            & 75.2 $\pm$ 0.8            & 77.2 $\pm$ 1.1            \\
DRO                 & 61.6 $\pm$ 0.7            & 52.9 $\pm$ 0.2            & 75.5 $\pm$ 0.5            & 77.7 $\pm$ 0.2            \\
Mixup                     & 64.7 $\pm$ 0.7            & 54.7 $\pm$ 0.6            & 77.3 $\pm$ 0.3            & 79.2 $\pm$ 0.3            \\
MLDG                      & 63.7 $\pm$ 0.3            & 54.5 $\pm$ 0.6            & 75.9 $\pm$ 0.4            & 78.6 $\pm$ 0.1            \\
CORAL                     & 64.4 $\pm$ 0.3            & 55.3 $\pm$ 0.5            & 76.7 $\pm$ 0.5            & 77.9 $\pm$ 0.5            \\
MMD                       & 63.0 $\pm$ 0.1            & 53.7 $\pm$ 0.9            & 76.1 $\pm$ 0.3            & 78.1 $\pm$ 0.5            \\
DANN                       & 59.3 $\pm$ 1.1            & 51.7 $\pm$ 0.2            & 74.1 $\pm$ 0.8            & 76.6 $\pm$ 0.6            \\
C-DANN                   & 61.0 $\pm$ 1.4            & 51.1 $\pm$ 0.7            & 74.1 $\pm$ 0.3            & 76.0 $\pm$ 0.7            \\
\bottomrule
\end{tabular}
\end{center}

\begin{center}
\begin{tabular}{lcccc}
\toprule
\multicolumn{5}{c}{\textbf{Model selection method: leave-one-domain-out cross-validation}} \\
\midrule
\textbf{Algorithm}    & \textbf{A}                & \textbf{C}                & \textbf{P}                & \textbf{R}                \\
\midrule
ERM                       & 62.3 $\pm$ 0.5            & 54.1 $\pm$ 0.5            & 75.3 $\pm$ 0.2            & 77.4 $\pm$ 0.5            \\
IRM                       & 62.1 $\pm$ 0.9            & 51.4 $\pm$ 0.6            & 75.5 $\pm$ 0.7            & 77.6 $\pm$ 0.8            \\
DRO                 & 62.7 $\pm$ 0.7            & 52.8 $\pm$ 1.0            & 75.4 $\pm$ 0.1            & 77.7 $\pm$ 0.2            \\
Mixup                     & 63.8 $\pm$ 0.4            & 52.9 $\pm$ 0.4            & 77.3 $\pm$ 0.4            & 78.7 $\pm$ 0.4            \\
MLDG                      & 62.9 $\pm$ 0.5            & 53.5 $\pm$ 0.7            & 76.0 $\pm$ 0.4            & 77.9 $\pm$ 0.6            \\
CORAL                     & 64.4 $\pm$ 0.3            & 55.4 $\pm$ 0.1            & 76.2 $\pm$ 0.2            & 78.4 $\pm$ 0.4            \\
MMD                       & 62.2 $\pm$ 0.3            & 52.7 $\pm$ 1.0            & 75.5 $\pm$ 0.4            & 78.1 $\pm$ 0.3            \\
DANN                       & 61.1 $\pm$ 0.1            & 51.6 $\pm$ 0.5            & 73.6 $\pm$ 0.5            & 75.8 $\pm$ 0.3            \\
C-DANN                   & 60.0 $\pm$ 0.4            & 50.2 $\pm$ 0.7            & 72.1 $\pm$ 1.0            & 76.4 $\pm$ 0.5            \\
\bottomrule
\end{tabular}
\end{center}

\begin{center}
\begin{tabular}{lcccc}
\toprule
\multicolumn{5}{c}{\textbf{Model selection method: test-domain validation set \textit{(oracle)}}} \\
\midrule
\textbf{Algorithm}    & \textbf{A}                & \textbf{C}                & \textbf{P}                & \textbf{R}                \\
\midrule
ERM                       & 61.2 $\pm$ 1.4            & 54.0 $\pm$ 0.5            & 75.9 $\pm$ 0.7            & 77.3 $\pm$ 0.4            \\
IRM                       & 62.4 $\pm$ 0.9            & 53.4 $\pm$ 0.7            & 75.5 $\pm$ 0.8            & 77.7 $\pm$ 0.6            \\
DRO                 & 63.6 $\pm$ 0.5            & 54.4 $\pm$ 0.7            & 75.9 $\pm$ 0.1            & 77.0 $\pm$ 0.4            \\
Mixup                     & 65.1 $\pm$ 0.6            & 54.6 $\pm$ 0.7            & 76.8 $\pm$ 0.6            & 77.7 $\pm$ 0.6            \\
MLDG                      & 61.0 $\pm$ 0.9            & 54.3 $\pm$ 0.3            & 75.8 $\pm$ 0.5            & 78.6 $\pm$ 0.1            \\
CORAL                     & 65.0 $\pm$ 0.5            & 54.3 $\pm$ 0.8            & 76.8 $\pm$ 0.4            & 78.2 $\pm$ 0.2            \\
MMD                       & 62.4 $\pm$ 0.2            & 53.6 $\pm$ 0.5            & 75.8 $\pm$ 0.4            & 76.4 $\pm$ 0.3            \\
DANN                       & 60.5 $\pm$ 0.9            & 51.9 $\pm$ 0.4            & 73.7 $\pm$ 0.4            & 76.4 $\pm$ 0.6            \\
C-DANN                   & 60.0 $\pm$ 0.5            & 52.0 $\pm$ 0.4            & 74.2 $\pm$ 0.5            & 76.3 $\pm$ 0.4            \\
\bottomrule
\end{tabular}
\end{center}

\clearpage
\newpage
\subsection{TerraIncognita}
\begin{center}
\begin{tabular}{lcccc}
\toprule
\multicolumn{5}{c}{\textbf{Model selection method: training domain validation set}} \\
\midrule
\textbf{Algorithm}    & \textbf{L100}             & \textbf{L38}              & \textbf{L43}              & \textbf{L46}              \\
\midrule
ERM                       & 50.8 $\pm$ 1.8            & 42.5 $\pm$ 0.7            & 57.9 $\pm$ 0.6            & 37.6 $\pm$ 1.2            \\
IRM                       & 52.2 $\pm$ 3.1            & 43.4 $\pm$ 2.4            & 57.7 $\pm$ 1.5            & 38.1 $\pm$ 0.7            \\
DRO                 & 47.2 $\pm$ 1.6            & 40.1 $\pm$ 1.6            & 57.6 $\pm$ 0.9            & 43.0 $\pm$ 0.7            \\
Mixup                     & 60.6 $\pm$ 1.3            & 41.1 $\pm$ 1.8            & 58.5 $\pm$ 0.8            & 35.2 $\pm$ 1.1            \\
MLDG                      & 48.5 $\pm$ 3.3            & 42.8 $\pm$ 0.4            & 56.8 $\pm$ 0.9            & 36.3 $\pm$ 0.5            \\
CORAL                     & 48.6 $\pm$ 0.9            & 42.2 $\pm$ 3.5            & 55.9 $\pm$ 0.6            & 38.7 $\pm$ 0.7            \\
MMD                       & 52.2 $\pm$ 5.8            & 47.0 $\pm$ 0.6            & 57.8 $\pm$ 1.3            & 40.3 $\pm$ 0.5            \\
DANN                       & 49.0 $\pm$ 3.8            & 46.3 $\pm$ 1.7            & 57.6 $\pm$ 0.8            & 40.6 $\pm$ 1.7            \\
C-DANN                   & 49.5 $\pm$ 3.8            & 44.8 $\pm$ 1.0            & 57.3 $\pm$ 1.1            & 38.8 $\pm$ 1.7            \\
\bottomrule
\end{tabular}
\end{center}

\begin{center}
\begin{tabular}{lcccc}
\toprule
\multicolumn{5}{c}{\textbf{Model selection method: leave-one-domain-out cross-validation}} \\
\midrule
\textbf{Algorithm}    & \textbf{L100}             & \textbf{L38}              & \textbf{L43}              & \textbf{L46}              \\
\midrule
ERM                       & 47.5 $\pm$ 0.2            & 43.8 $\pm$ 0.2            & 55.4 $\pm$ 1.3            & 38.3 $\pm$ 1.3            \\
IRM                       & 44.2 $\pm$ 2.7            & 41.3 $\pm$ 0.6            & 54.3 $\pm$ 2.0            & 36.0 $\pm$ 1.7            \\
DRO                 & 31.8 $\pm$ 0.3            & 43.7 $\pm$ 1.2            & 58.0 $\pm$ 0.7            & 36.6 $\pm$ 1.3            \\
Mixup                     & 49.6 $\pm$ 4.8            & 44.4 $\pm$ 0.9            & 55.0 $\pm$ 1.4            & 35.2 $\pm$ 1.9            \\
MLDG                      & 50.9 $\pm$ 5.1            & 39.9 $\pm$ 0.9            & 58.0 $\pm$ 1.8            & 34.6 $\pm$ 1.0            \\
CORAL                     & 51.8 $\pm$ 2.2            & 42.1 $\pm$ 1.1            & 59.6 $\pm$ 0.8            & 38.7 $\pm$ 2.3            \\
MMD                       & 51.5 $\pm$ 1.7            & 37.4 $\pm$ 2.0            & 58.9 $\pm$ 0.9            & 37.4 $\pm$ 1.8            \\
DANN                       & 47.2 $\pm$ 4.5            & 40.6 $\pm$ 0.0            & 55.7 $\pm$ 2.6            & 39.4 $\pm$ 1.3            \\
C-DANN                   & 43.2 $\pm$ 3.5            & 30.9 $\pm$ 4.1            & 50.4 $\pm$ 4.4            & 37.8 $\pm$ 1.5            \\
\bottomrule
\end{tabular}
\end{center}

\begin{center}
\begin{tabular}{lcccc}
\toprule
\multicolumn{5}{c}{\textbf{Model selection method: test-domain validation set \textit{(oracle)}}} \\
\midrule
\textbf{Algorithm}    & \textbf{L100}             & \textbf{L38}              & \textbf{L43}              & \textbf{L46}              \\
\midrule
ERM                       & 59.9 $\pm$ 1.0            & 48.7 $\pm$ 0.4            & 58.9 $\pm$ 0.3            & 43.3 $\pm$ 0.9            \\
IRM                       & 56.8 $\pm$ 2.0            & 46.5 $\pm$ 0.3            & 57.9 $\pm$ 0.6            & 42.4 $\pm$ 0.5            \\
DRO                 & 61.2 $\pm$ 1.2            & 47.5 $\pm$ 0.6            & 59.5 $\pm$ 0.6            & 44.1 $\pm$ 0.8            \\
Mixup                     & 65.1 $\pm$ 1.8            & 46.8 $\pm$ 0.6            & 59.5 $\pm$ 0.3            & 40.0 $\pm$ 1.1            \\
MLDG                      & 58.7 $\pm$ 0.5            & 48.9 $\pm$ 0.7            & 59.5 $\pm$ 0.4            & 42.4 $\pm$ 0.6            \\
CORAL                     & 60.5 $\pm$ 1.0            & 47.6 $\pm$ 1.8            & 59.1 $\pm$ 0.3            & 43.2 $\pm$ 0.5            \\
MMD                       & 60.0 $\pm$ 1.6            & 46.7 $\pm$ 0.8            & 60.0 $\pm$ 1.0            & 44.2 $\pm$ 0.4            \\
DANN                       & 57.6 $\pm$ 1.3            & 48.1 $\pm$ 1.1            & 58.2 $\pm$ 0.5            & 42.7 $\pm$ 1.4            \\
C-DANN                   & 56.3 $\pm$ 2.6            & 46.9 $\pm$ 1.5            & 57.8 $\pm$ 0.8            & 43.3 $\pm$ 0.5            \\
\bottomrule
\end{tabular}
\end{center}

\clearpage
\newpage
\subsection{DomainNet}

\begin{center}
\begin{tabular}{lcccccc}
\toprule
\multicolumn{7}{c}{\textbf{Model selection method: training domain validation set}} \\
\midrule
\textbf{Algorithm}    & \textbf{clipart} & \textbf{infograph} & \textbf{painting} & \textbf{quickdraw} & \textbf{real} & \textbf{sketch}     \\
\midrule
ERM          &    58.4 $\pm$ 0.3 &  19.2 $\pm$ 0.4 &  46.3 $\pm$ 0.5 & 12.8 $\pm$ 0.0 & 60.6 $\pm$ 0.5 & 49.7 $\pm$ 0.8 \\            
IRM          &    51.0 $\pm$ 3.3 &  16.8 $\pm$ 1.0 &  38.8 $\pm$ 2.1 & 11.8 $\pm$ 0.5 & 51.5 $\pm$ 3.6 & 44.2 $\pm$ 3.1 \\            
DRO    &    47.8 $\pm$ 0.6 &  17.1 $\pm$ 0.6 &  36.6 $\pm$ 0.7 & 8.8 $\pm$ 0.4  & 51.5 $\pm$ 0.6 & 40.7 $\pm$ 0.3 \\            
Mixup        &    55.3 $\pm$ 0.3 &  18.2 $\pm$ 0.3 &  45.0 $\pm$ 1.0 & 12.5 $\pm$ 0.3 & 57.1 $\pm$ 1.2 & 49.2 $\pm$ 0.3 \\            
MLDG         &    59.5 $\pm$ 0.0 &  19.8 $\pm$ 0.4 &  48.3 $\pm$ 0.5 & 13.0 $\pm$ 0.4 & 59.5 $\pm$ 1.0 & 50.4 $\pm$ 0.7 \\            
CORAL        &    58.7 $\pm$ 0.2 &  20.9 $\pm$ 0.3 &  47.3 $\pm$ 0.3 & 13.6 $\pm$ 0.3 & 60.2 $\pm$ 0.3 & 50.2 $\pm$ 0.6 \\            
MMD          &    54.6 $\pm$ 1.7 &  19.3 $\pm$ 0.3 &  44.9 $\pm$ 1.1 & 11.4 $\pm$ 0.5 & 59.5 $\pm$ 0.2 & 47.0 $\pm$ 1.6 \\            
DANN          &    53.8 $\pm$ 0.7 &  17.8 $\pm$ 0.3 &  43.5 $\pm$ 0.3 & 11.9 $\pm$ 0.5 & 56.4 $\pm$ 0.3 & 46.7 $\pm$ 0.5 \\            
C-DANN      &    53.4 $\pm$ 0.4 &  18.3 $\pm$ 0.7 &  44.8 $\pm$ 0.3 & 12.9 $\pm$ 0.2 & 57.5 $\pm$ 0.4 & 46.7 $\pm$ 0.2 \\            
\bottomrule
\end{tabular}
\end{center}

\begin{center}
\begin{tabular}{lcccccc}
\toprule
\multicolumn{7}{c}{\textbf{Model selection method: leave-one-domain-out cross-validation}} \\
\midrule
\textbf{Algorithm}    & \textbf{clipart} & \textbf{infograph} & \textbf{painting} & \textbf{quickdraw} & \textbf{real} & \textbf{sketch}     \\
\midrule
ERM                 &      56.0 $\pm$ 1.1     &       19.6 $\pm$ 0.2  &          47.3 $\pm$ 0.3    &        12.5 $\pm$ 0.3      &      60.5 $\pm$ 0.5     &       49.1 $\pm$ 0.2 \\            
IRM                 &      49.0 $\pm$ 2.4     &       16.7 $\pm$ 0.9  &          38.8 $\pm$ 2.1    &        10.2 $\pm$ 0.6      &      53.2 $\pm$ 2.1     &       43.7 $\pm$ 2.1 \\            
DRO           &      47.3 $\pm$ 0.7     &       16.8 $\pm$ 0.3  &          35.2 $\pm$ 0.1    &        8.8 $\pm$ 0.4       &      50.1 $\pm$ 2.3     &       38.9 $\pm$ 0.7 \\            
Mixup               &      54.4 $\pm$ 0.6     &       18.1 $\pm$ 0.3  &          45.2 $\pm$ 0.3    &        12.1 $\pm$ 0.4      &      57.9 $\pm$ 1.1     &       48.6 $\pm$ 0.1 \\            
MLDG                &      58.7 $\pm$ 0.4     &       20.3 $\pm$ 0.1  &          48.8 $\pm$ 0.1    &        13.0 $\pm$ 0.4      &      61.2 $\pm$ 0.2     &       50.3 $\pm$ 0.2 \\            
CORAL               &      57.9 $\pm$ 0.7     &       20.8 $\pm$ 0.3  &          47.5 $\pm$ 0.4    &        13.5 $\pm$ 0.3      &      61.0 $\pm$ 0.3     &       50.6 $\pm$ 0.5 \\            
MMD                 &      54.0 $\pm$ 2.2     &       19.3 $\pm$ 0.3  &          44.9 $\pm$ 1.1    &        11.4 $\pm$ 0.5      &      59.5 $\pm$ 0.2     &       47.0 $\pm$ 1.6 \\            
DANN                 &      53.1 $\pm$ 0.4     &       17.5 $\pm$ 0.6  &          42.8 $\pm$ 0.4    &        10.2 $\pm$ 0.5      &      56.4 $\pm$ 0.3     &       44.9 $\pm$ 0.9 \\            
C-DANN             &      53.4 $\pm$ 0.4     &       18.3 $\pm$ 0.7  &          44.2 $\pm$ 0.5    &        12.9 $\pm$ 0.2      &      57.1 $\pm$ 0.2     &       46.7 $\pm$ 0.2 \\            
\bottomrule
\end{tabular}
\end{center}

\begin{center}
\begin{tabular}{lcccccc}
\toprule
\multicolumn{7}{c}{\textbf{Model selection method: test-domain validation set \textit{(oracle)}}} \\
\midrule
\textbf{Algorithm}    & \textbf{clipart} & \textbf{infograph} & \textbf{painting} & \textbf{quickdraw} & \textbf{real} & \textbf{sketch}     \\
\midrule
ERM               &         58.4 $\pm$ 0.3        &     19.8 $\pm$ 0.2     &        47.3 $\pm$ 0.3      &       13.4 $\pm$ 0.2      &       60.7 $\pm$ 0.5      &       49.9 $\pm$ 0.7 \\            
IRM               &         51.0 $\pm$ 3.3        &     16.7 $\pm$ 0.9     &        38.8 $\pm$ 2.1      &       11.8 $\pm$ 0.5      &       53.2 $\pm$ 2.1      &       44.7 $\pm$ 2.7 \\            
DRO         &         47.8 $\pm$ 0.6        &     17.2 $\pm$ 0.6     &        36.3 $\pm$ 0.5     &        9.0 $\pm$ 0.2      &        52.8 $\pm$ 0.3      &       40.7 $\pm$ 0.3 \\            
Mixup             &         55.8 $\pm$ 0.6        &     19.2 $\pm$ 0.2     &        46.2 $\pm$ 0.6      &       12.8 $\pm$ 0.2      &       58.7 $\pm$ 0.6      &       49.2 $\pm$ 0.3 \\            
MLDG              &         59.3 $\pm$ 0.2        &     20.3 $\pm$ 0.1     &        48.8 $\pm$ 0.1      &       14.0 $\pm$ 0.3      &       61.2 $\pm$ 0.2      &       51.2 $\pm$ 0.1 \\            
CORAL             &         58.8 $\pm$ 0.1        &     20.8 $\pm$ 0.3     &        47.5 $\pm$ 0.4      &       13.6 $\pm$ 0.2      &       61.0 $\pm$ 0.3      &       50.8 $\pm$ 0.4 \\            
MMD               &         54.6 $\pm$ 1.7        &     19.6 $\pm$ 0.1     &        44.9 $\pm$ 1.1      &       12.6 $\pm$ 0.1      &       59.7 $\pm$ 0.2      &       47.5 $\pm$ 1.2 \\            
DANN               &         53.8 $\pm$ 0.7        &     17.5 $\pm$ 0.6     &        43.5 $\pm$ 0.3      &       11.8 $\pm$ 0.6      &       56.4 $\pm$ 0.3      &       46.7 $\pm$ 0.5 \\            
C-DANN           &         53.4 $\pm$ 0.4        &     18.4 $\pm$ 0.6     &        44.7 $\pm$ 0.3      &       12.9 $\pm$ 0.2      &       57.5 $\pm$ 0.4      &       46.5 $\pm$ 0.2 \\            
\bottomrule
\end{tabular}
\end{center}

\clearpage
\newpage

\section{Dataset details}
\label{sec:dataset_details}

\domainbed includes downloaders and loaders for seven multi-domain image classification tasks: 

\begin{itemize}
    \item \textbf{Colored MNIST} \citep{arjovsky2019invariant} is a variant of the MNIST handwritten digit classification dataset \citep{lecun1998mnist}. Domain $d \in \{0.1, 0.3, 0.9\}$ contains a disjoint set of digits colored either red or blue. The label is a noisy function of the digit and color, such that color bears correlation $d$ with the label and the digit bears correlation 0.75 with the label. This dataset contains $70,000$ examples of dimension $(2, 28, 28)$ and $2$ classes.
    \item \textbf{Rotated MNIST} \citep{ghifary2015domain} is a variant of MNIST where domain $d \in \{$ 0, 15, 30, 45, 60, 75 $\}$ contains digits rotated by $d$ degrees. Our dataset contains $70,000$ examples of dimension $(1, 28, 28)$ and $10$ classes.
    \item \textbf{PACS} \citep{Li_2017_ICCV} comprises four domains $d \in \{$ \text{art}, \text{cartoons}, \text{photos}, \text{sketches} $\}$. This dataset contains $9,991$ examples of dimension $(3, 224, 224)$ and $7$ classes.
    \item \textbf{VLCS} \citep{fang2013unbiased} comprises photographic domains $d \in \{$ \text{Caltech101}, \text{LabelMe}, \text{SUN09}, \text{VOC2007} $\}$. This dataset contains $10,729$ examples of dimension $(3, 224, 224)$ and $5$ classes.
    \item \textbf{Office-Home} \citep{venkateswara2017Deep} includes domains $d \in \{$ \text{art}, \text{clipart}, \text{product}, \text{real} $\}$. This dataset contains $15,588$ examples of dimension $(3, 224, 224)$ and $65$ classes.
    \item \textbf{Terra Incognita} \citep{beery2018recognition} contains photographs of wild animals taken by camera traps at locations $d \in \{ \text{L100}, \text{L38}, \text{L43}, \text{L46}\}$. Our version of this dataset contains $24,788$ examples of dimension $(3, 224, 224)$ and $10$ classes.
    \item \textbf{DomainNet} \citep{peng2019moment} has six domains $d \in \{$ \text{clipart}, \text{infograph}, \text{painting}, \text{quickdraw}, \text{real}, \text{sketch} $\}$. This dataset contains $586,575$ examples of size $(3, 224, 224)$ and $345$ classes.
\end{itemize}

For all datasets, we first pool the raw training, validation, and testing images together.
For each random seed, we then instantiate random training, validation, and testing splits.

\clearpage
\newpage
\section{Model architectures, hyperparameter spaces, and other training details}
\label{sec:architectures}

In this section we describe the model architectures and hyperparameter search spaces used in our experiments.

\subsection{Architectures}

We list the neural network architecture used for each dataset in Table \ref{table:architectures} and specify the details of our MNIST network in \ref{table:mnist_convnet}.

\begin{minipage}{0.45\textwidth}
    \begin{table}[H]
    \caption{Neural network architectures used for each dataset.} 
        \begin{tabular}{ll}
        \toprule
        \textbf{Dataset} & \textbf{Architecture} \\
        \midrule
        Colored MNIST & \multirow{2}{*}{MNIST ConvNet} \\
        Rotated MNIST & \\
        \midrule
        PACS & \multirow{5}{*}{ResNet-50} \\
        VLCS & \\
        Office-Home &  \\
        TerraIncognita &  \\
        DomainNet &  \\
        \bottomrule
        \end{tabular}
\label{table:architectures}
\end{table}
\end{minipage}
\hfill
\begin{minipage}{0.48\textwidth}
\begin{table}[H]
    \caption{Details of our MNIST ConvNet architecture. All convolutions use 3 $\times$ 3 kernels and ``same'' padding.} 
        \begin{tabular}{ll}
        \toprule
        \textbf{\#} & \textbf{Layer}\\
        \midrule
            1  & Conv2D (in=$d$, out=64)\\
            2  & ReLU\\
            3  & GroupNorm (groups=8)\\
            4  & Conv2D (in=64, out=128, stride=2)\\
            5  & ReLU\\
            6  & GroupNorm (8 groups)\\
            7  & Conv2D (in=128, out=128)\\
            8  & ReLU\\
            9  & GroupNorm (8 groups)\\
            10 & Conv2D (in=128, out=128)\\
            11 & ReLU\\
            12 & GroupNorm (8 groups)\\
            13 & Global average-pooling\\
        \bottomrule
        \end{tabular}
\label{table:mnist_convnet}
\end{table}
\end{minipage}

For the architecture ``Resnet-50'', we replace the final (softmax) layer of a ResNet50 pretrained on ImageNet and fine-tune. 
Observing that batch normalization interferes with domain generalization algorithms (as different minibatches follow different distributions), we freeze all batch normalization layers before fine-tuning.
We insert a dropout layer before the final linear layer.

\clearpage
\newpage
\subsection{Hyperparameters}

We list all hyperparameters, their default values, and the search distribution for each hyperparameter in our random hyperparameter sweeps, in Table \ref{table:hyperparameters}.

\begin{table}[H]
    \caption{Hyperparameters, their default values and distributions for random search.} 
    \begin{center}
    { 
    \begin{tabular}{llll}
        \toprule
        \textbf{Condition} & \textbf{Parameter} & \textbf{Default value} & \textbf{Random distribution}\\
        \midrule
        \multirow{4}{*}{ResNet}       & learning rate & 0.00005 & $10^{\text{Uniform}(-5, -3.5)}$\\
                                      & batch size    & 32   & $2^{\text{Uniform}(3, 5.5)}$\\
                                      & generator learning rate & 0.00005 & $10^{\text{Uniform}(-5, -3.5)}$\\
                                      & discriminator learning rate & 0.00005 & $10^{\text{Uniform}(-5, -3.5)}$\\
        \midrule
        \multirow{4}{*}{not ResNet}   & learning rate & 0.001 & $10^{\text{Uniform}(-4.5, -3.5)}$\\
                                      & batch size    & 64   & $2^{\text{Uniform}(3, 9)}$\\
                                      & generator learning rate & 0.001 & $10^{\text{Uniform}(-4.5, -2.5)}$\\
                                      & discriminator learning rate & 0.001 & $10^{\text{Uniform}(-4.5, -2.5)}$\\
        \midrule
        \multirow{2}{*}{MNIST}        & weight decay & 0    & 0\\
                                      & generator weight decay & 0    & 0\\
        \midrule
        \multirow{2}{*}{not MNIST}    & weight decay & 0    & $10^{\text{Uniform}(-6, -2)}$\\
                                      & generator weight decay & 0    & $10^{\text{Uniform}(-6, -2)}$\\
        \midrule
        \multirow{5}{1.5cm}{DANN, C-DANN} & lambda                 & 1.0    & $10^{\text{Uniform}(-2, 2)}$\\
         & discriminator weight decay & 0    & $10^{\text{Uniform}(-6, -2)}$\\
         & discriminator steps        & 1    & $2^{\text{Uniform}(0, 3)}$\\
         & gradient penalty           & 0    & $10^{\text{Uniform}(-2, 1)}$\\
         & adam $\beta_1$             & 0.5    & $\text{RandomChoice}([0, 0.5])$\\
        \midrule
        \multirow{2}{*}{IRM}          & lambda  & 100    & $10^{\text{Uniform}(-1, 5)}$\\
                                      & iterations of penalty annealing & 500 & $10^{\text{Uniform}(0, 4)}$\\
        \midrule
        Mixup                         & alpha & 0.2 & $10^{\text{Uniform}(0, 4)}$\\
        \midrule
        DRO                           & eta   & 0.01 & $10^{\text{Uniform}(-1, 1)}$\\
        \midrule
        MMD                           & gamma & 1 & $10^{\text{Uniform}(-1, 1)}$\\
        \midrule
        MLDG           & beta & 1 & $10^{\text{Uniform}(-1, 1)}$\\
        \midrule
        all          & dropout & 0    & $\text{RandomChoice}([0, 0.1, 0.5])$\\
        \bottomrule
    \end{tabular}
    }
    \end{center}
    \label{table:hyperparameters}
\end{table}

\subsection{Other training details}

We optimize all models using Adam~\citep{kingma2014adam}.

\clearpage
\newpage
\section{Adding new datasets and algorithms to our framework}
\label{sec:code_example}

In their basic form, algorithms are classes that implement a method \texttt{.update(minibatches)} and a method \texttt{.predict(x)}.
The update method receives a list of minibatches, one minibatch per training domain, and each minibatch containing a number of input-output pairs. 
For example, to implement group DRO \citep[Algorithm 1]{sagawa2019distributionally}, we simply write the following in \texttt{algorithms.py}:

\lstset{
    language=Python,
    basicstyle=\ttfamily,
    keywordstyle=\color{blue}\ttfamily,
    stringstyle=\color{red}\ttfamily,
    commentstyle=\color{green}\ttfamily,
    morecomment=[l][\color{magenta}]{\#},
}

\newsavebox{\lsta}
\begin{lrbox}{\lsta}
\begin{lstlisting}
class DRO(ERM):
    def __init__(self, input_shape, num_classes, num_domains, hparams):
        super().__init__(input_shape, num_classes, num_domains, hparams)
        self.register_buffer("q", torch.Tensor())

    def update(self, minibatches):
        device = "cuda" if minibatches[0][0].is_cuda else "cpu"

        if not len(self.q):
            self.q = torch.ones(len(minibatches)).to(device)

        losses = torch.zeros(len(minibatches)).to(device)

        for m in range(len(minibatches)):
            x, y = minibatches[m]
            losses[m] = F.cross_entropy(self.predict(x), y)
            self.q[m] *= (self.hparams["dro_eta"] * losses[m].data).exp()

        self.q /= self.q.sum()
        loss = torch.dot(losses, self.q) / len(minibatches)

        self.optimizer.zero_grad()
        loss.backward()
        self.optimizer.step()

        return {'loss': loss.item()}
\end{lstlisting}
\end{lrbox}

\scalebox{0.75}{\usebox{\lsta}}

By inheriting from ERM, this new class has access to a default classifier \texttt{.network}, optimizer \texttt{.optimizer}, and prediction method \texttt{.predict(x)}.
Finally, we should tell \domainbed about the hyperparameters of this new algorithm. To do so, add the following line to the function \texttt{\_hparams} from \texttt{hparams\_registry.py}:

\newsavebox{\lstc}
\begin{lrbox}{\lstc}
\begin{lstlisting}
hparams['dro_eta'] = (1e-2, 10**random_state.uniform(-3, -1))
\end{lstlisting}
\end{lrbox}

\scalebox{0.75}{\usebox{\lstc}}

To add a new image classification dataset to \domainbed, arrange your image files as \texttt{/path/MyDataset/domain/class/image.jpg}.
Then, append to \texttt{datasets.py}:

\newsavebox{\lstb}
\begin{lrbox}{\lstb}
\begin{lstlisting}
class MyDataset(MultipleEnvironmentImageFolder):
    N_STEPS = 2500
    CHECKPOINT_FREQ = 300
    def __init__(self, root, test_envs=None):
        self.dir = os.path.join(root, "MyDataset/")
        super(MyDataset, self).__init__(self.dir)
\end{lstlisting}
\end{lrbox}

\scalebox{0.75}{\usebox{\lstb}}

In the previous, \texttt{N\_STEPS} determines the number of gradient updates an algorithm should perform to learn this dataset.
The variable \texttt{CHECKPOINT\_FREQ} determines the number of gradient steps an algorithm should wait before reporting its performance in all domains.

We are now ready to launch an experiment with our new algorithm and dataset:

\newsavebox{\lstd}
\begin{lrbox}{\lstd}
\begin{lstlisting}
python train.py --model DRO --dataset MyDataset --data_dir /path --test\_envs 1 \
                --output_dir /path/to/logs_files --hparams '{"dro_eta": 0.2}'
\end{lstlisting}
\end{lrbox}

\scalebox{0.75}{\usebox{\lstd}}

Finally, we can run a fully automated sweep on all datasets, algorithms, test domains, and model selection criteria by simply invoking \texttt{python sweep.py}. After adapting the file \texttt{sweep.py} to the computing infrastructure at hand, this single command automatically generates all the result tables that we report in this manuscript. 

\subsection{Extension to UDA}

By extending the method \texttt{.update(minibatches, unlabeled)} to accept a minibatch of unlabeled examples from the test domain, we can immediately use \domainbed as a framework to perform experimentation on unsupervised domain adaptation algorithms.

\end{document}